\documentclass[letterpaper]{article} 
\usepackage{aaai23}  
\usepackage{times}  
\usepackage{helvet}  
\usepackage{courier}  
\usepackage[hyphens]{url}  
\usepackage{graphicx} 
\usepackage{natbib}  
\usepackage{caption} 
\usepackage{algorithm}
\usepackage{algorithmic}
\usepackage[draft]{hyperref} 

\usepackage[utf8]{inputenc} 
\usepackage[T1]{fontenc}    
\usepackage{booktabs}       
\usepackage{amsfonts}       
\usepackage{nicefrac}       
\usepackage{microtype}      
\usepackage{xcolor}        
\usepackage{tikz}
\usepackage{pgfplots}
\usepackage{xcolor}
\usepackage{amssymb}
\usepackage{wrapfig}
\usepackage{lipsum}

\usepackage{caption}
\usepackage{subcaption}

\usepackage{filecontents}
\usepackage{algorithm}
\usepackage{algorithmic}
\usepackage{algorithm}
\usepackage{wrapfig}
\usepackage{lipsum}
\usepackage{amsmath}
\usepackage{graphicx}
\usepackage{url}
\usepackage{amsfonts}
\usepackage{color, soul}
\usepackage{mathrsfs}
\usepackage{cleveref}
\usepackage{multirow}
\usepackage{float}
\usepackage{wrapfig}
\usepackage{amsthm}
\usepackage{bm}
\usepackage{bbm}
\usepackage{algorithm}
\usepackage{algorithmic}
\usepackage{colortbl}
\usepackage{enumitem}

\newtheorem{myDef}{Definition}
\newtheorem{myDef1}{Problem}
\newtheorem{myDef2}{Proposition}
\newtheorem{myDef3}{Theorem}
\newtheorem{myDef4}{Corollary}

\usepackage{multirow}
\usepackage{newfloat}
\usepackage{listings}
\DeclareCaptionStyle{ruled}{labelfont=normalfont,labelsep=colon,strut=off} 
\floatstyle{ruled}
\newfloat{listing}{tb}{lst}{}
\floatname{listing}{Listing}
\title{Interpreting Unfairness in Graph Neural Networks via Training Node Attribution}

\author{
    Yushun Dong\textsuperscript{\rm 1}, Song Wang\textsuperscript{\rm 1}, Jing Ma\textsuperscript{\rm 1}, Ninghao Liu\textsuperscript{\rm 2}, Jundong Li\textsuperscript{\rm 1}
}
\affiliations{
    \textsuperscript{\rm 1}University of Virginia, 
    \textsuperscript{\rm 2}University of Georgia\\

    \{yd6eb, sw3wv, jm3mr, jundong\}@virginia.edu, ninghao.liu@uga.edu
}

\usepackage{bibentry}

\begin{document}

\maketitle

\begin{abstract}

Graph Neural Networks (GNNs) have emerged as the leading paradigm for solving graph analytical problems in various real-world applications. Nevertheless, GNNs could potentially render biased predictions towards certain demographic subgroups. 
Understanding how the bias in predictions arises is critical, as it guides the design of GNN debiasing mechanisms.
However, most existing works overwhelmingly focus on GNN debiasing, but fall short on explaining how such bias is induced.
In this paper, we study a novel problem of interpreting GNN unfairness through attributing it to the influence of training nodes.
Specifically, we propose a novel strategy named Probabilistic Distribution Disparity (PDD) to measure the bias exhibited in GNNs, and develop an algorithm to efficiently estimate the influence of each training node on such bias.
We verify the validity of PDD and the effectiveness of influence estimation through experiments on real-world datasets.
Finally, we also demonstrate how the proposed framework could be used for debiasing GNNs. Open-source code can be found at https://github.com/yushundong/BIND.
\end{abstract}

\section{Introduction}
\label{intro}

Graph data is pervasive among a plethora of realms, e.g., financial fraud detection~\cite{wang2019semi,pourhabibi2020fraud,cheng2020graph}, social recommendation~\cite{fan2019graph,song2019session,guo2020deep}, and chemical reaction prediction~\cite{do2019graph,shi2020graph,kwon2022uncertainty}. As one of the state-of-the-art approaches to handle graph data, Graph Neural Networks (GNNs) have been attracting increasing attention~\cite{DBLP:conf/iclr/KipfW17,hamilton2017inductive,velivckovic2017graph}. 
Over the years, various graph analytical tasks have benefited from GNNs, where node classification is among the most widely studied ones~\cite{DBLP:conf/iclr/KipfW17,wu2019simplifying,wu2020comprehensive}. 
%
%
Nevertheless, in node classification, GNNs often yield results with discrimination towards specific demographic subgroups described by certain sensitive attributes~\cite{dong2021edits,dai2021say,agarwal2021towards,zhang2022fairness,wang2022improving}, such as gender, race, and religion.
In many high-stake applications, critical decisions are made based on the classification results of GNNs~\cite{shumovskaia2020linking}, e.g., crime forecasting~\cite{jin2020addressing}, and the exhibited bias (i.e., unfairness) is destructive for the involved individuals~\cite{dong2022fairness,dong2022structural,song2022guide}.
To tackle this problem, there has been a line of works focusing on debiasing GNNs in node classification~\cite{dong2021edits,dai2021say,agarwal2021towards,dong2021individual,loveland2022fairedit,dai2022learning}.
Their goal is to relieve the bias in GNN predictions on the test set and in this paper we refer to it as model bias.


%
%
In addition to debiasing GNNs, it is also critical to interpret how the model bias arises in GNNs. This is because such an understanding not only helps to determine whether a specific node should be involved in the training set, but also has much potential to guide the design of GNN debiasing methods~\cite{dong2021edits,loveland2022fairedit,li2020dyadic}.
Nevertheless, most existing GNN interpretation methods aim to understand how a prediction is made~\cite{yuan2020explainability,Liu2022} instead of other aspects such as fairness.
Consequently, although the graph data has been proved to be a significant source of model bias~\cite{dong2021edits,li2020dyadic}, existing works are unequipped to tackle this problem.
In this paper, we aim to address this problem at the instance (node) level. Specifically, given a GNN trained for node classification, we aim to answer: ``\textit{To what extent the GNN model bias is influenced by the existence of a specific training node in this graph?}''

%


Nevertheless, answering the above question is technically challenging. Essentially, there are three main challenges: 
(1) \emph{Influence Quantification.}
To depict the influence of each training node on the model bias of GNNs, the first and foremost challenge is to design a principled fairness metric. A straightforward approach is to directly employ traditional fairness metrics (e.g., $\Delta_{\text{SP}}$ for \textit{Statistical Parity}~\cite{dwork2012fairness} and $\Delta_{\text{EO}}$ for \textit{Equal Opportunity}~\cite{DBLP:conf/nips/HardtPNS16}). 
However, these metrics are not applicable in our task.
The reason is that most of them are computed based on the predicted labels, while a single training node can barely twist these predicted labels on test data~\cite{zhang2022dotin,sun2020adversarial}.
%
%
Consequently, the influence of a single training node on the model bias would be hard to capture. 
(2) \emph{Computation Efficiency.} 
To compute the influence of each training node on the model bias, a natural way is to re-train the GNN on a new graph with this specific training node being deleted and observe how the exhibited model bias changes.
%
%
However, such a re-training process is prohibitively expensive.
%
%
%
%
%
%
%
(3) \emph{Non-I.I.D. Characterization.} Graph data goes against the widely adopted i.i.d. assumption, as neighboring nodes are often dependent on each other~\cite{ma2021subgroup,ying2019gnnexplainer}. 
Therefore, when a specific node is deleted from the graph, all its neighbors could exert different influences on the model bias of GNN during training.
%
%
Such complex dependencies bring obstacles towards the analysis of node influence on the model bias.

To tackle the above challenges, in this paper, we propose a novel framework named 
BIND (\underline{B}iased tra\underline{I}ning \underline{N}ode i\underline{D}entification)
to quantify and estimate the influence of each training node on the model bias of GNNs.
Specifically, to handle the first challenge, we propose \textit{Probabilistic Distribution Disparity} (PDD) as a principled strategy to quantify the model bias.
PDD directly quantifies the exhibited bias in the GNN probabilistic output instead of the predicted labels.
Therefore, PDD is with finer granularity and is more suitable for capturing the influence of each specific training node compared with traditional fairness metrics.
%
%
%
To handle the second challenge, we propose an estimation algorithm for the node influence on model bias, which avoids the re-training process and thus achieves better efficiency.
%
%
To tackle the third challenge, we also characterize the dependency between nodes based on the analysis of the training loss for GNNs.
Finally, experiments on real-world datasets corroborate the effectiveness of BIND. Our contributions are mainly summarized as (1) \textbf{Problem Formulation.} We formulate a novel problem of interpreting the bias exhibited in GNNs through attributing to the influence of training nodes; (2) \textbf{Metric and Algorithm Design.} 
%
We propose a novel framework BIND to quantify and efficiently estimate the influence of each training node on the model bias of GNNs; (3) \textbf{Experimental Evaluation.} We perform comprehensive experiments on real-world datasets to evaluate the effectiveness of the proposed framework BIND.

\section{Preliminaries}

We first present the notations used in this paper. Then, we define the problem of interpreting GNN unfairness through quantifying the influence of each specific training node.

\subsection{Notations}
\label{notations}

In this paper, matrices, vectors, and scalars are represented with bold uppercase letters (e.g., $\bm{A}$), bold lowercase letters (e.g., $\bm{x}$), and normal lowercase letters (e.g., $n$), respectively. 

We denote an input graph as $\mathcal{G} = \{\mathcal{V}, \mathcal{E}, \mathcal{X}\}$, where $\mathcal{V} = \{v_1, ..., v_n\}$ denotes the node set, $\mathcal{E} \subseteq \mathcal{V} \times \mathcal{V}$ represents the edge set, $\mathcal{X} = \{\bm{x}_1, ..., \bm{x}_n\}$ is the node attribute vectors, and $\bm{x}_i$ ($1 \leq i \leq n$) represents the attribute vector of node $v_i$.
We denote $\mathcal{G}_{-i}$ as the new graph with node $v_i$ being deleted from $\mathcal{G}$.
Additionally, we employ $\mathcal{V}'$ ($\mathcal{V}' \subseteq \mathcal{V}$) to represent the training node set, where $|\mathcal{V}'| = m$.
%
The nodes in graph $\mathcal{G}$ are mapped to the output space with a trained GNN $f_{\bm{W}}$, where $\bm{W}$ represents the learnable parameters of the GNN model.
We denote the optimized parameters (i.e., the parameters after training) as $\hat{\bm{W}}$.
In node classification, the probabilistic classification output for the $n$ nodes is denoted as $\mathcal{\hat{Y}} = \{\hat{\bm{y}}_1, ..., \hat{\bm{y}}_n\}$, where $\hat{\bm{y}}_i \in \mathbb{R}^{c}$, and $c$ is the number of classes.
%
We use $Y$ and $S$ to denote the ground truth label and the sensitive attribute for nodes, respectively.
%
%
%
For an $L$-layer GNN $f_{\bm{W}}$, we define the subgraph up to $L$ hops away centered on $v_i$ as its computation graph (denoted as $\mathcal{G}_i = \{\mathcal{V}_i, \mathcal{E}_i, \mathcal{X}_i\}$).
Here $\mathcal{V}_i, \mathcal{E}_i,$ and $\mathcal{X}_i$ denote the set of nodes, edges, and node attributes in $\mathcal{G}_i$, respectively.
It is worth noting that existing works have proven that $\mathcal{G}_i$ fully determines the information $f_{\bm{W}}$ utilizes to make the prediction of $v_i$~\cite{ying2019gnnexplainer}.
%
%
For node $v_i$, we use $\mathcal{V}_i'$ to indicate the intersection between $\mathcal{V}_i$ and $\mathcal{V}'$, i.e., $\mathcal{V}_i' = \mathcal{V}_i \cap \mathcal{V}'$, which is the set of training nodes in $\mathcal{G}_i$.

\subsection{Problem Statement}


The problem of interpreting GNN unfairness is formally defined as follows.

\begin{myDef1}
\label{p1}
\textbf{GNN Unfairness Interpretation.} Given the graph data $\mathcal{G}$ and a GNN model $f_{\hat{\bm{W}}}$ trained based on $\mathcal{G}$, we define the problem of interpreting GNN unfairness as to quantify the influence of each training node to the unfairness exhibited in GNN predictions on the test set.
\end{myDef1}

\section{Methodology}

In this section, we first briefly introduce GNNs for the node classification task. 
Then, to tackle the challenge of \emph{Influence Quantification}, we propose Probabilistic Distribution Disparity (PDD) to measure model bias and define node influence on the bias in a trained GNN.
%
%
%
%
Furthermore, to tackle the challenge of \emph{Computation Efficiency}, we design an algorithm to estimate the node influence on the model bias.
Finally, we introduce how to characterize the dependency between nodes in influence estimation, which tackles the challenge of \emph{Non-I.I.D. Characterization}.

\subsection{GNNs in Node Classification}

In the node classification task, GNNs take the input graph $\mathcal{G}$ and output a probabilistic output matrix $\hat{\bm{Y}}$, where the $i$-th row in $\hat{\bm{Y}}$ is $\hat{\bm{y}}_i$, i.e., the probabilistic prediction of a node's membership over all possible classes.
Usually, there are multiple layers in GNNs, where the formulation of the $l$-th layer can be summarized as:
\begin{align}
\!\!\bm{z}_{i}^{(l+1)}\!=\!\sigma\left(\operatorname{AGG}\left(\bm{z}_{i}^{(l)}, h \left(\left\{\bm{z}_{j}^{(l)}: v_j \in \mathcal{N}(v_i)\right\}\right)\right)\right).
\end{align}
Here $\bm{z}_{i}^{(l)}$ is the embedding of node $i$ at the $l$-th layer; $\mathcal{N}(v_i)$ is the set of one-hop neighbors around $v_i$; $h(\cdot)$ is a function with learnable parameters; $\operatorname{AGG}(\cdot)$ and $\sigma(\cdot)$ denote the aggregation function (e.g., mean operator) and activation function (e.g., ReLU), respectively.
Later on, a loss function $L_{\mathcal{V}'}$ (e.g., cross-entropy loss) defined on the set of training nodes $\mathcal{V}'$ is employed for GNN training.

\subsection{Probabilistic Distribution Disparity}

%


Traditional bias metrics such as $\Delta_{\text{SP}}$ for statistical parity and $\Delta_{\text{EO}}$ for equal opportunity are computed on the predicted class labels. However, a single training node can hardly twist these predicted labels~\cite{zhang2022dotin,sun2020adversarial}. Hence the node-level contribution to model bias can barely be captured by traditional bias metrics.
To capture the influence of a single training node on model bias, we propose Probabilistic Distribution Disparity (PDD) as a novel bias quantification strategy.
PDD can be instantiated with different fairness notions to depict the model bias from different perspectives.
%
%
%
%
Specifically, we assume the population is divided into different sensitive subgroups, i.e., demographic subgroups described by the sensitive attribute.
To achieve finer granularity, we define PDD as the Wasserstein-1 distance~\cite{kantorovich1960mathematical} between the probability distributions of a variable of interest in different sensitive subgroups. Compared with traditional fairness metrics, continuous changes brought by each specific training node are reflected in the measured distributions, and Wasserstein distance is theoretically more sensitive to the change of the measured distributions over other commonly used distribution distance metrics~\cite{ArjovskyCB17}. In addition, we note that the variable of interest depends on the chosen fairness notion in applications, and a larger value of PDD indicates a higher level of model bias.
%
%
We introduce two instantiations of PDD based on two traditional fairness notions, including \textit{Statistical Parity}~\cite{dwork2012fairness} and \textit{Equal Opportunity}~\cite{DBLP:conf/nips/HardtPNS16}. Both notions are based on binary classification tasks and binary sensitive attributes (generalizations to non-binary cases can be found in Appendix A).  
%
For example, Statistical Parity requires the probability of positive predictions to be the same across two sensitive subgroups, where the variable of interest is the GNN probabilistic output $\hat{\bm{y}}$. 
%
We use $\mathcal{\hat{Y}}^{(S=j)}$ to denote the set of the probabilistic predictions for test nodes whose sensitive attribute $S$ equals to $j$ ($j \in \{0, 1\}$).
%
Let the distribution of the probabilistic predictions in $\mathcal{\hat{Y}}^{(S=0)}$ and $\mathcal{\hat{Y}}^{(S=1)}$ be $P^{(S=0)}_{\hat{\bm{y}}}$ and $P^{(S=1)}_{\hat{\bm{y}}}$, respectively.
The PDD instantiated with statistical parity $\Gamma_{SP}$ is
\begin{align}
\label{fair_cost1}  
    \Gamma_{SP} = \text{Wasserstein}_{1} (P^{(S=0)}_{\hat{\bm{y}}}, P^{(S=1)}_{\hat{\bm{y}}}),
\end{align}
where $\text{Wasserstein}_1( \cdot , \cdot )$ takes two distributions as input and outputs the Wasserstein-1 distance between them.
Denote $Y$ as the ground truth for node classification.
Similarly, we can also instantiate PDD based on Equal Opportunity $\Gamma_{EO}$ as
\begin{align}
\label{fair_cost2}  
    \Gamma_{EO} = \text{Wasserstein}_1 (P^{(S=0, Y=1)}_{\hat{\bm{y}}}, P^{(S=1, Y=1)}_{\hat{\bm{y}}}).
\end{align}
$P^{(S=0, Y=1)}_{\hat{\bm{y}}}$ and $P^{(S=1, Y=1)}_{\hat{\bm{y}}}$ are model prediction distributions for nodes with $(S=0, Y=1)$ and $(S=1, Y=1)$, respectively.
%
%
With such a strategy, we then define node influence on model bias.

\begin{myDef}
\label{d1}
\textbf{Node Influence on Model Bias.} 
Let $f_{\hat{\bm{W}}}$ and $f_{\hat{\bm{W}}'}$ denote the GNN model trained on graph $\mathcal{G}$ and $\mathcal{G}_{-i}$ (i.e., $\mathcal{G}$ with node $v_i\in \mathcal{V}'$ being deleted), respectively.
Let $\Gamma_1$ and $\Gamma_2$ be the Probabilistic Distribution Disparity value based on the output of $f_{\hat{\bm{W}}}$ and $f_{\hat{\bm{W}}'}$ for nodes in test set. We define $\Delta \Gamma = \Gamma_2 - \Gamma_1$ as the influence of node $v_i$ on the model bias.
\end{myDef}
The rationale behind this definition is to measure to what extent $\Gamma$ changes if the GNN model is trained on a graph without $v_i$.
Thus, $\Delta \Gamma$ depicts the influence of node $v_i$ on the model bias. 
%
For both instantiations of $\Gamma$ (i.e., $\Gamma_{SP}$ and $\Gamma_{EO}$), if $\Delta \Gamma >0$, deleting the training node $v_i$ from $\mathcal{G}$ leads to a more unfair (or biased) GNN model. This indicates that node $v_i$ contributes to improving the fairness level, i.e., $v_i$ is helpful for fairness. Nevertheless, the above computation requires re-training the GNN to obtain the influence of each training node, which is too expensive if we want to compute the influence of all nodes in the training set.
%
%
In Section~\ref{estimation_methodology}, we introduce how to efficiently estimate $\Delta \Gamma$.

\subsection{Node Influence on Model Bias Estimation}
\label{estimation_methodology}

\begin{figure*}[!t]
    \vspace{-2mm}
    \centering
    \includegraphics[width=0.7\textwidth]{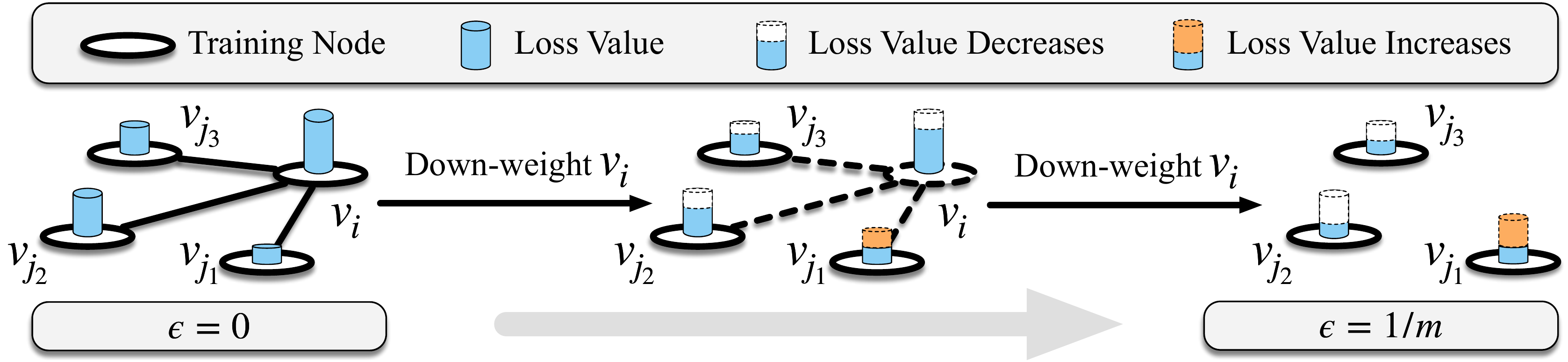}
    \vspace{-1mm}
    \caption{An illustration of how down-weighting node $v_i$ influences the loss values of the training nodes in $\mathcal{G}_i$ (including $v_i$, $v_{j_1}$, $v_{j_2}$, and $v_{j_3}$). Scenarios from $\epsilon=0$ to $\epsilon=1/m$ are presented.}
    \vspace{-5mm}
    \label{illustration}
\end{figure*} 


It is noteworthy that PDD is a function of $\hat{\bm{W}}$ for a trained GNN, as $\hat{\bm{W}}$ directly determines the probabilistic predictions for test nodes.
Hence we first characterize how a training node in $\mathcal{G}$ influences $\hat{\bm{W}}$, followed by how this node influences PDD via applying the chain rule.
%
%
Formally, the optimal parameters $\hat{\bm{W}}$ minimize the objective function $L_{\mathcal{V}'}(\mathcal{G}, \bm{W})$ of the node classification task, so that:
\begin{align}
\hat{\bm{W}} \stackrel{\text { def }}{=} \arg \min_{\bm{W}} L_{\mathcal{V}'}(\mathcal{G}, \bm{W}) \notag  = \arg \min_{\bm{W}} \frac{1}{m} \sum_{i=1}^{m} L_{v_i}\left(\mathcal{G}_i, \bm{W}\right).
\end{align}
%
Here $L_{v_i}\left(\mathcal{G}_i, \bm{W}\right)$ denotes the loss term associated with node $v_i$; $\mathcal{G}_i$ is the computation graph of $v_i$; $m$ is the total number of training nodes.
%
%
If a training node $v_i$ is deleted from $\mathcal{G}$, the loss function will change and thus leads to a different $\hat{\bm{W}}$.
%
We take $v_i$ as an example to analyze the influence on $\hat{\bm{W}}$ after deleting a training node from $\mathcal{G}$.
%
Traditionally, the existence of node $v_i$ is considered as a binary state, which is either one (if $v_i$ exists in $\mathcal{G}$) or zero (otherwise). But in our analysis, we treat it as a continuous variable to depict the intermediate states of the existence of $v_i$.
Suppose that the existence of $v_i$ is down-weighted in the training of a GNN on $\mathcal{G}$. This operation leads to two changes in the loss function: 
(1) the loss term associated with node $v_i$, i.e., $L_{v_i}\left(\mathcal{G}_i, \bm{W}\right)$, is down-weighted;
(2) the loss terms associated with other training nodes in the computation graph of $v_i$ would also be influenced. The reason is that these nodes could be affected by the information from node $v_i$ during the message passing in GNNs~\cite{DBLP:conf/iclr/KipfW17,ying2019gnnexplainer}. %
Based on the above analysis, 
%
%
%
%
%
%
we define $\hat{\bm{W}}_{\epsilon, v_i}$ as the optimal parameter that minimizes the loss function when node $v_i$ is down-weighted as follows:
%
%
\begin{align}
\label{change}
\hat{\bm{W}}_{\epsilon, v_i} \stackrel{\text { def }}{=} &\arg \min_{\bm{W}}  L_{\mathcal{V}'}(\mathcal{G}, \bm{W}) \notag \\ &- \epsilon \left(L_{v_i}\left(\mathcal{G}_i, \bm{W}\right) + \tilde{L}_{\mathcal{V}_i'}(\mathcal{G}_i, \bm{W})\right),
\end{align}
%
where $\epsilon \in [0, 1/m]$ 
controls the scale of down-weighting $v_i$.
An illustration in Fig.~\ref{illustration} shows how down-weighting $v_i$ affects the loss values of training nodes in its computation graph.
%
%
To formally characterize how node $v_i$ influences $\bm{\hat{W}}$, we have Theorem~\ref{I1} as follows (see proofs in Appendix C). 
\begin{myDef3}
\label{I1} 
According to the optimization objective of $\bm{\hat{W}}_{\epsilon, v_{i}}$ in Eq.~(\ref{change}), we have
\begin{small}
\begin{align}
\label{influence1} 
\left.  \frac{d \bm{\hat{W}}_{\epsilon, v_{i}}}{d \epsilon}\right|_{\epsilon=0}=  &\left(\frac{\partial^{2} L_{\mathcal{V}'}(\mathcal{G},\hat{\bm{W}})}{\partial\bm{W}^{2}} \right)^{-1}   \notag \\ & \cdot \left(\frac{\partial L_{v_i}\left(\mathcal{G}_i,\hat{\bm{W}}\right)}{\partial\bm{W}} + \frac{\partial \tilde{L}_{\mathcal{V}_i'}(\mathcal{G}_i,\hat{\bm{W}})}{\partial\bm{W}}\right).
\end{align}
\end{small}
\end{myDef3}
Then, we characterize the influence of down-weighting node $v_i$ on the value of PDD.
We present Corollary~\ref{I_G} based on the chain rule as follows (see the proofs in Appendix C).  
\begin{myDef4}
\label{I_G} 
Define the derivative of $\Gamma$ w.r.t. $\epsilon$ at $\epsilon=0$ as $I_{\Gamma}(v_i)$.
%
%
According to Theorem~\ref{I1}, we have
\begin{align}
\label{nabla}
\left. I_{\Gamma}(v_i) \stackrel{\text { def }}{=}  \frac{\partial \Gamma}{\partial \epsilon}  \right|_{\epsilon=0} =  \left(\frac{\partial \Gamma}{\partial \bm{W}} \right)^{\top} \left.  \frac{d \bm{\hat{W}}_{\epsilon, v_{i}}}{d \epsilon}\right|_{\epsilon=0}.
\end{align}
\end{myDef4}
%
%
With Corollary~\ref{I_G}, we can now estimate the value change of $\Gamma$ when node $v_i$ is down-weighted via
\begin{align}
\label{value_change}
\Gamma_{\epsilon,v_i} - \Gamma_{0,v_i} =  - \epsilon \cdot I_{\Gamma}(v_i) + o(\epsilon) \approx  - \epsilon \cdot I_{\Gamma}(v_i)
\end{align}
according to the first-order Taylor expansion.
%
Here $\Gamma_{\epsilon,v_i}$ and $\Gamma_{0,v_i}$ are the PDD values after and before node $v_i$ is down-weighted, respectively.
To estimate the value change in $\Gamma$ for a GNN trained on $\mathcal{G}_{-i}$,
we introduce Theorem~\ref{t1} as follows (see the proofs in Appendix C).
\begin{myDef3}
\label{t1} 
Compared with the GNN trained on $\mathcal{G}$, $\Delta \Gamma = \Gamma_{\frac{1}{m},v_i} - \Gamma_{0,v_i}$ is equivalent to the value change in $\Gamma$ when the GNN mode is trained on graph $\mathcal{G}_{-i}$.
\end{myDef3}
Theorem~\ref{t1} enables us to directly compute the $\Delta \Gamma$ for an arbitrary training node $v_i$, which helps avoid the expensive re-training process.
In the next section, we further define $\tilde{L}_{\mathcal{V}_i'}(\mathcal{G}_i, \hat{\bm{W}})$ and present an algorithm to efficiently estimate the node influence on model bias.

\subsection{Non-I.I.D. Characterization}
\label{non-iid}


Generally, there are two types of dependencies between a training node $v_i$ and other nodes in its computation graph $\mathcal{G}_i$, namely its dependency on other training nodes and its dependency on test nodes.
The dependency between training nodes directly influences $\bm{W}$ during GNN training, and thus influences the probabilistic outcome of all test nodes. Hence it is critical to properly characterize the dependency between $v_i$ and other training nodes.
%
%
Specifically, we aim to characterize how the loss summation of all training nodes in $\mathcal{G}_i$ changes due to the existence of $v_i$.
%
%
We denote the training nodes other than node $v_i$ in $\mathcal{G}_i$ as $\mathcal{V}_i' \backslash \{v_i\}$.
%
%
For any node $v_j \in \mathcal{V}_i' \backslash \{v_i\}$, we denote $\mathcal{G}_{j,-i}$ as the computation graph of node $v_j$ with node $v_i$ being deleted.
$\tilde{L}_{\mathcal{V}_i}(\mathcal{G}_i,\hat{\bm{W}})$ is then formally given as
\begin{small}
\begin{align}
\label{delta_l_modeling}
\tilde{L}_{\mathcal{V}_i}(\mathcal{G}_i,\hat{\bm{W}}) = \sum_{v_j \in \mathcal{V}_i' \backslash \{v_i\}} \left( L_{v_j}\left(\mathcal{G}_j,\hat{\bm{W}}\right) -  L_{v_j}\left(\mathcal{G}_{j,-i}, \hat{\bm{W}}\right) \right).
\end{align}  
\end{small}
The first term represents the summation of loss for nodes in $\mathcal{V}_i' \backslash \{v_i\}$ on $\mathcal{G}$, and the second term denotes the summation of loss for these nodes on $\mathcal{G}_{-i}$. In this regard, $\tilde{L}_{\mathcal{V}_i}(\mathcal{G}_i,\hat{\bm{W}})$ generally depicts to what extent the loss summation changes for nodes in $\mathcal{V}_i' \backslash \{v_i\}$ on graph $\mathcal{G}$ compared with $\mathcal{G}_{-i}$.
%
%
If $v_i$ is down-weighted by a certain degree, the change of the loss summation for nodes in $\mathcal{V}_i' \backslash \{v_i\}$ can be depicted by a linearly re-scaled $\tilde{L}_{\mathcal{V}_i}(\mathcal{G}_i,\hat{\bm{W}})$, as described in Eq.~(\ref{change}).
%

Additionally, there could also be dependencies between $v_i$ and test nodes in $\mathcal{G}_i$, as $v_i$ can influence the representations of its neighboring test nodes due to the information propagation mechanism in GNNs during inference.
Such a dependency could also influence the value of PDD when $v_i$ is deleted from $\mathcal{G}$.
Correspondingly, we introduce the characterization of the dependency between $v_i$ and test nodes.
Specifically, we 
%
%
%
present an upper bound to depict the normalized change magnitude of the neighboring test nodes' representations when a training node $v_i$ is deleted.
Here the analysis is based on the prevalent GCN model~\cite{DBLP:conf/iclr/KipfW17}, and can be easily generalized to other GNNs.
Following widely adopted assumptions in~\cite{huang2020graph,xu2018representation}, we have Proposition~\ref{non-training} (see the proofs in Appendix C).



\begin{algorithm}[H]
  \footnotesize
    \caption{Node Influence on Model Bias Estimation}
    \begin{algorithmic}[1]
    \label{algorithm1}
      \REQUIRE 
$\mathcal{G}$: the graph data; 
$f_{\hat{\bm{W}}}$: the trained GNN model; 
$\mathcal{V}'$: the set of training nodes;
\ENSURE 
$\mathcal{I}_{\Gamma} = \{\Gamma_{\frac{1}{m},v_i} - \Gamma_{0,v_i}: v_i \in \mathcal{V}'\}$; \\
\STATE Initialize $\mathcal{I}_{\Gamma} = \varnothing$; 
\STATE Compute $\{\frac{\partial \Gamma}{\partial \bm{W}} : v_i \in \mathcal{V}'\}$ based on $f_{\hat{\bm{W}}}$; 
\WHILE{$v_i \in \mathcal{V}'$}
\STATE Compute $\left.  \frac{d \bm{\hat{W}}_{\epsilon, v_{i}}}{d \epsilon}\right|_{\epsilon=0}$ according to Eq.~(\ref{influence1}) and (\ref{delta_l_modeling}); 
\STATE Compute $I_{\Gamma}(v_i)$ according to Eq.~(\ref{nabla}); 
\STATE Compute $\Gamma_{\frac{1}{m},v_i} - \Gamma_{0,v_i}$ according to Eq.~(\ref{value_change}); 
\STATE Append element $\Gamma_{\frac{1}{m},v_i} - \Gamma_{0,v_i}$ onto $\mathcal{I}_{\Gamma}$; 
\ENDWHILE
\RETURN $\mathcal{I}_{\Gamma}$;
\end{algorithmic}
\end{algorithm}


\begin{myDef2}
\label{non-training} 
Denote the representations of node $v_j (v_j \in \mathcal{V} \backslash \mathcal{V}')$ based on $\mathcal{G}$ and $\mathcal{G}_{-i}$ as $\bm{z}_j$ and $\bm{z}_j^{\star}$, respectively.
Define $h^{(j,i)}$ and $q^{(j,i)}$ as the distance from $v_j$ to $v_i$ and the number of all possible paths from $v_j$ to $v_i$, respectively.
Define the set of geometric mean node degrees of $q^{(j,i)}$ paths as $\mathcal{D} = \{d_{1}^{(j,i)}, ..., d_{q^{(j,i)}}^{(j,i)}\}$.
Define $d_{min}^{(j,i)}$ as the minimum value of $\mathcal{D}$.
Assume the norms of all node representations are the same.
We then have $\|\bm{z}_j^{\star}-\bm{z}_j\|_2/\|\bm{z}_j\|_2\leq q^{(j,i)}/(d_{min}^{(j,i)})^{h^{(j,i)}}$.
\end{myDef2}

From Proposition~\ref{non-training}, we observe that (1) deleting $v_i$ exerts an upper-bounded impact on the representations of other test nodes in its computation graph; and (2) this upper-bound exponentially decays w.r.t. the distance between $v_i$ and test nodes.
Hence the dependency between $v_i$ and test nodes has limited influence on $\Gamma$ during inference when $v_i$ is deleted from the graph.
%
%
On the contrary, considering that the dependency between $v_i$ and other training nodes directly influences $\hat{\bm{W}}$ and thus influences the inference results of all nodes, such a dependency should not be neglected.
Consequently, we argue that it is reasonable to estimate the influence of each training node on $\Gamma$ by only considering the dependency between training nodes.
We present the algorithmic routine of $\Delta \Gamma$ estimation in Algorithm~\ref{algorithm1}.

\subsection{Complexity Analysis}

To better understand the computational cost, here we analyze the time complexity of estimating $\Delta \Gamma$ according to Algorithm~\ref{algorithm1}.
We denote the number of parameters in $\bm{W}$ and the average number of training nodes in the computation graph of an arbitrary training node as $t$ and $\bar{r}$, respectively.
For each node $v_i$, the time complexity to compute $\partial L_{v_i}\left(\mathcal{G}_i,\hat{\bm{W}}\right) / \partial\bm{W}$ and $\partial \tilde{L}_{\mathcal{V}_i'}(\mathcal{G}_i,\hat{\bm{W}}) / \partial\bm{W}$ is $O$($t$) and $O$($\bar{r}t$), respectively.
Hence the time complexity is $O$($m\bar{r}t$) to traverse all training nodes.
For the Hessian matrix inverse, we employ a widely-used estimation approach (see Appendix A) with linear time complexity w.r.t $t$.
Thus the time complexity of Eq.~(\ref{influence1}) and (\ref{delta_l_modeling}) is $O$($m\bar{r}t$).
Additionally, the time complexity of Eq.~(\ref{nabla}) and (\ref{value_change}) is $O$($mt$) and $O$($m$), respectively.
%
%
To summarize, the time complexity of Algorithm~\ref{algorithm1} is $O$($m\bar{r}t$). Considering that $\bar{r} \leq m$, the algorithm has a quadratic time complexity w.r.t. training node number.
This verifies the impressive time efficiency of our algorithm.
%


\section{Experiments}

We aim to answer the following research questions in experiments.
\textbf{RQ1}: How efficient is BIND in estimating the influence of training nodes on the mode bias?
\textbf{RQ2}: How well can BIND estimate the influence of training nodes on the model bias?
\textbf{RQ3}: How well can we debias GNNs via deleting harmful training nodes based on our estimation?
More details of experimental settings, supplementary experiments, and further analysis are in Appendix B.

\subsection{Experimental Setup}
\label{experimental_settings}

\noindent \textbf{Downstream Task \& Datasets.}
Here the downstream task is node classification. Four real-world datasets are adopted in our experiments, including \textit{Income}, \textit{Recidivism}, \textit{Pokec-z}, and \textit{Pokec-n}.
Specifically, \textit{Income} is collected from \textit{Adult Data Set}~\cite{Dua:2019}. Each individual is represented by a node, and we establish connections (i.e., edges) between individuals following a similar criterion adopted in~\cite{agarwal2021towards}. The sensitive attribute is race, and the task is to classify whether the salary of a person is over \$50K per year or not. 
\textit{Recidivism} is collected from~\cite{jordan2015effect}. A node represents a defendant released on bail, and defendants are connected based on their similarity. The sensitive attribute is race, and the task is to classify whether a defendant is on bail or not.
\textit{Pokec-z} and \textit{Pokec-n} are collected from \textit{Pokec}, which is a popular social network in Slovakia~\cite{takac2012data}.
In both datasets, each user is a node, and each edge stands for the friendship relation between two users. Here the locating region of users is the sensitive attribute. The task is to classify the working field of users. 
More details are in Appendix B.


\noindent \textbf{Baselines \& GNN Backbones.}
We compare our method with three state-of-the-art GNN debiasing baselines, namely FairGNN~\cite{dai2021say}, NIFTY~\cite{agarwal2021towards}, and EDITS~\cite{dong2021edits}.
To perform GNN debiasing,
FairGNN employs adversarial training to filter out the information of sensitive attributes from node embeddings;
NIFTY maximizes the agreement between the predictions based on perturbed sensitive attributes and unperturbed ones;
EDITS pre-processes the input graph data to be less biased via attribute and structural debiasing.
%
%
%
We mainly present the results of using GCN~\cite{DBLP:conf/iclr/KipfW17} as the backbone GNN model, while experiments with other GNNs are discussed in Appendix B.

\noindent \textbf{Evaluation Metrics.}
First, we employ running speedup factors to evaluate efficiency.
Second, we use the widely adopted Pearson Correlation~\cite{koh2017understanding,chen2020multi} between the estimated and actual $\Delta \Gamma$ to evaluate the effectiveness of node influence estimation.
Third, we adopt two traditional fairness metrics, namely $\Delta_{\text{SP}}$ (the metric for \textit{Statistical Parity})~\cite{dwork2012fairness} and $\Delta_{\text{EO}}$ (the metric for \textit{Equal Opportunity})~\cite{hardt2016equality}, to evaluate the effectiveness of debiasing GNNs via harmful nodes deletion. Additionally, the classification accuracy is also employed to evaluate the utility-fairness trade-off.

\begin{figure}[t!]
     \centering
     \includegraphics[width=.35\textwidth]{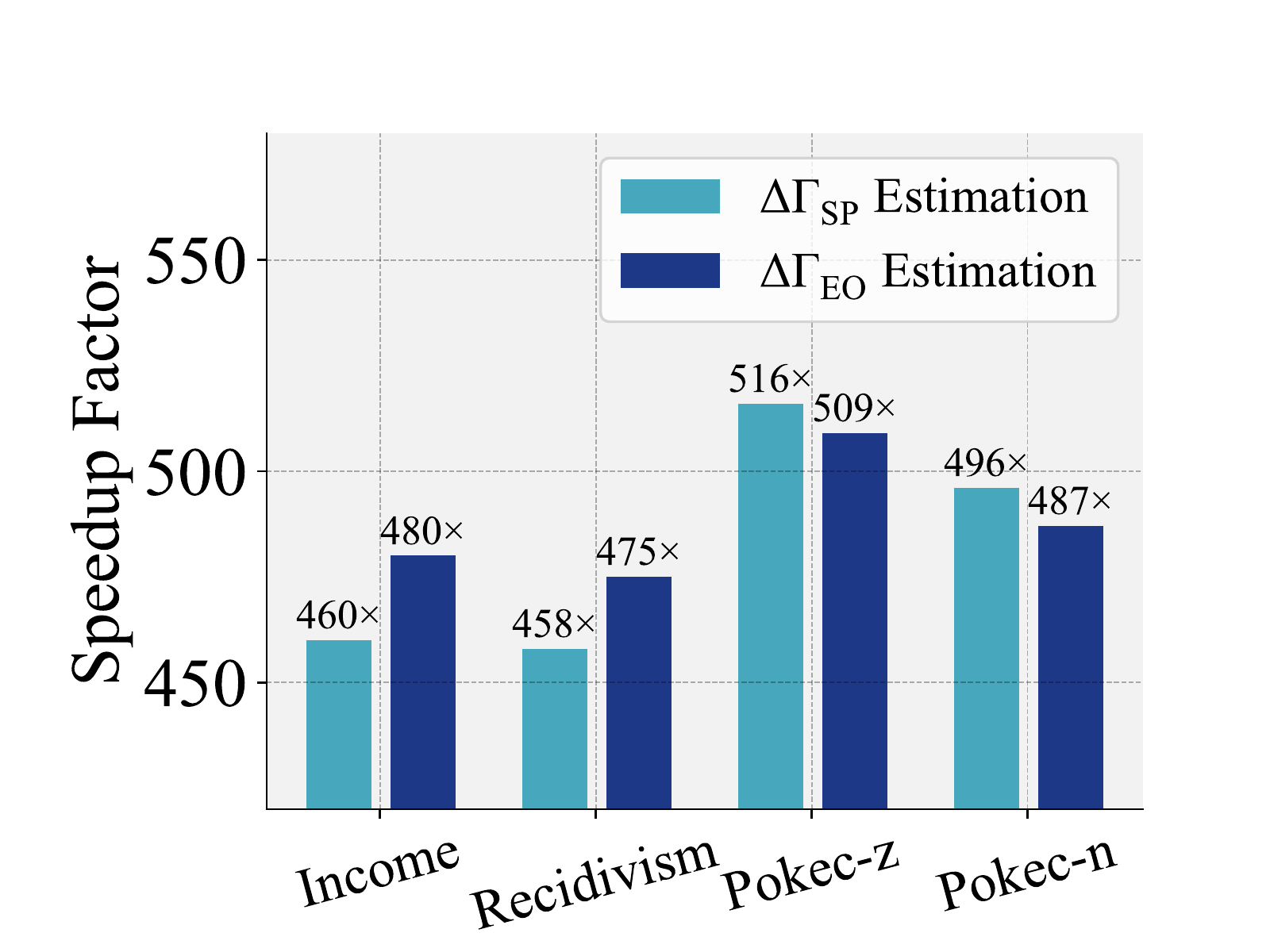}
\vspace{-2mm}
     \caption{Evaluation of efficiency: speedup factors of $\Delta \Gamma_{\text{SP}}$ and $\Delta \Gamma_{\text{EO}}$ estimation over GNN re-training.}
     \label{speedup}
\vspace{-2mm}
\end{figure}


\subsection{Efficiency of Node Influence Estimation}  

To answer RQ1, we evaluate the efficiency of $\Delta \Gamma$ estimation by comparing its running time with that of GNN re-training.
The running time of GNN re-training is computed as follows. We first delete the target node from the original input graph $\mathcal{G}$ and re-train the GCN to obtain $f_{\hat{\bm{W}}'}$.
We then obtain $\Delta \Gamma$ based on the values of $\Gamma$ given by $f_{\hat{\bm{W}}}$ and $f_{\hat{\bm{W}}'}$.
The above running time is defined as the time cost of GNN re-training.
The running time averaged across all training nodes is compared between GNN re-training and BIND, and we present the running speedup factors of BIND on the four real-world datasets in Fig.~\ref{speedup}.
We observe that the running speedup factors are over 450$\times$ on all four real-world datasets, which corroborates the efficiency superiority of BIND in estimating the value of $\Delta \Gamma$.
Additionally, we observe that the estimation on Pokec-z and Pokec-n datasets has higher speedup factors on both $\Delta \Gamma_{\text{SP}}$ and $\Delta \Gamma_{\text{EO}}$ compared with the other two datasets.
A reason could be that nodes in Pokec-z and Pokec-n have lower average degrees (see Appendix B). This facilitates the computation of $\tilde{L}_{\mathcal{V}_i'}(\mathcal{G}_i, \hat{\bm{W}})$ (the term that characterizes non-i.i.d.) and corresponding derivatives.

\begin{figure}[t!]
     \centering
     \includegraphics[width=.35\textwidth]{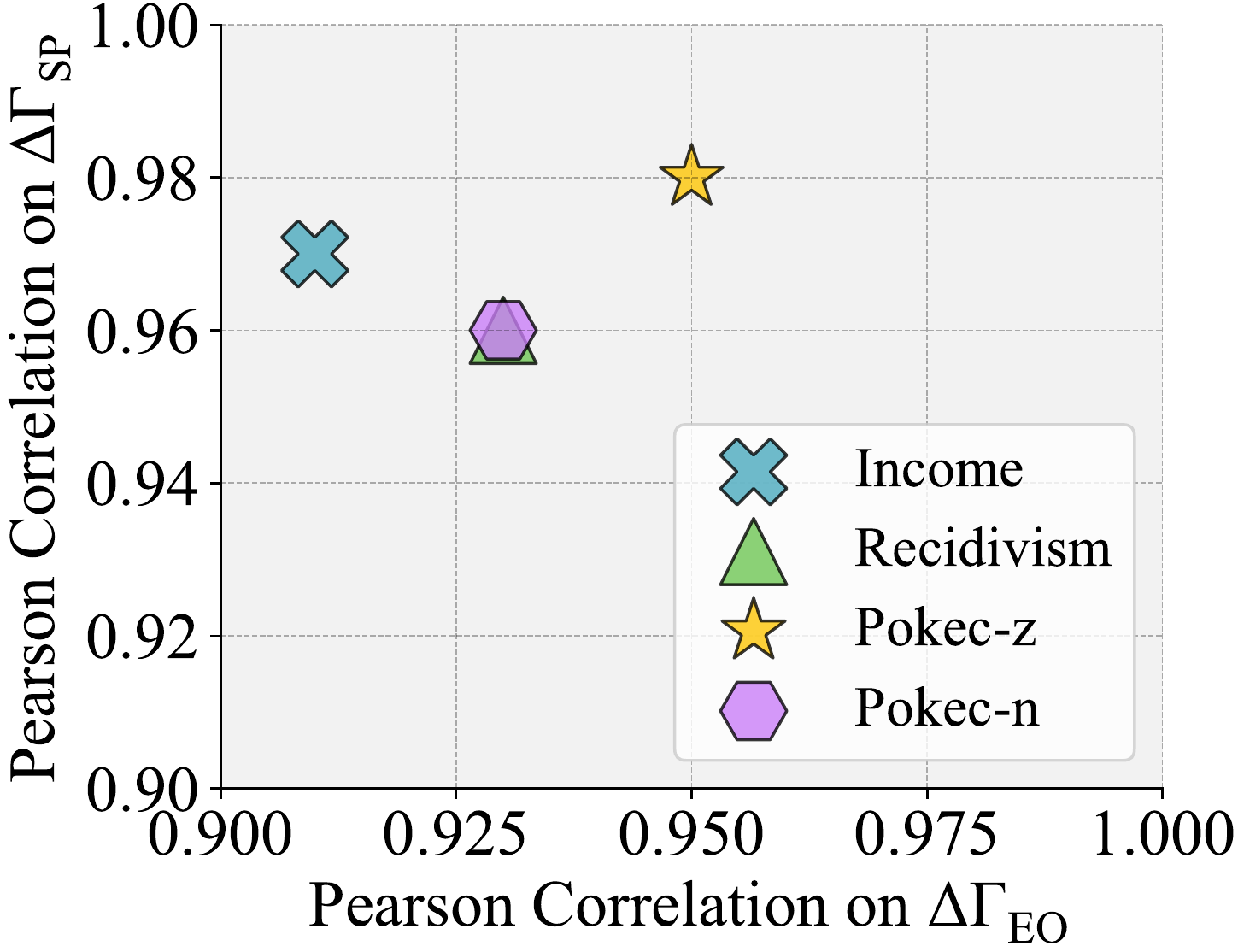}
     \caption{Evaluation of effectiveness: correlation between estimated and actual $\Delta \Gamma_{\text{SP}}$ and $\Delta \Gamma_{\text{EO}}$.}
     \label{points}
\vspace{-4mm}
\end{figure}


\subsection{Effectiveness of Node Influence Estimation}  %

\label{effectiveness}

We now evaluate the effectiveness of $\Delta \Gamma$ estimation.
%
It is worth noting that the numerical values of the estimated influence on model bias are small for most of the nodes (see Appendix B).
Here we introduce a strategy to evaluate the estimation effectiveness across a wider value range of $\Delta \Gamma$.
The basic intuition here is that we select node sets and evaluate how well their estimated $\Delta \Gamma$ summation aligns with the actual one.
Specifically, we first follow the widely adopted routine~\cite{koh2017understanding,chen2020multi} to truncate the helpful and harmful nodes with top-ranked $\Delta \Gamma$ values.
We then construct a series of node sets associated with the largest positive and negative estimated $\Delta \Gamma$ summations under different set size thresholds. The range of these thresholds is between zero and a maximum possible value (determined by the training set size).
%
%
It is worth noting that only nodes with non-overlapping computation graphs are selected in constructing each node set. This ensures that these nodes result in an estimated $\Delta \Gamma$ equivalent to the summation of their estimated $\Delta \Gamma$ (see Appendix C).
%
%
%
%
We present the Pearson correlation of estimated $\Delta \Gamma_{\text{SP}}$ and $\Delta \Gamma_{\text{EO}}$ with the actual values on four datasets in Fig.~\ref{points}.
It is worth noting that achieving an exact linear correlation (i.e., Pearson correlation equals one) between the estimated and actual $\Delta \Gamma$ is almost impossible, since we only employ the first-order Taylor expansion in our estimation for $\Delta \Gamma$.
From Fig.~\ref{points}, we observe that the estimation achieves Pearson correlation values over 0.9 on both $\Gamma_{\text{SP}}$ and $\Gamma_{\text{EO}}$ across all datasets.
Such consistencies between estimated and actual values verify the effectiveness of BIND.

\begin{figure*}[!t]
\centering
    \vspace{1mm}
        \begin{subfigure}[t]{0.21\textwidth}
        \small
        \includegraphics[width=1.04\textwidth]{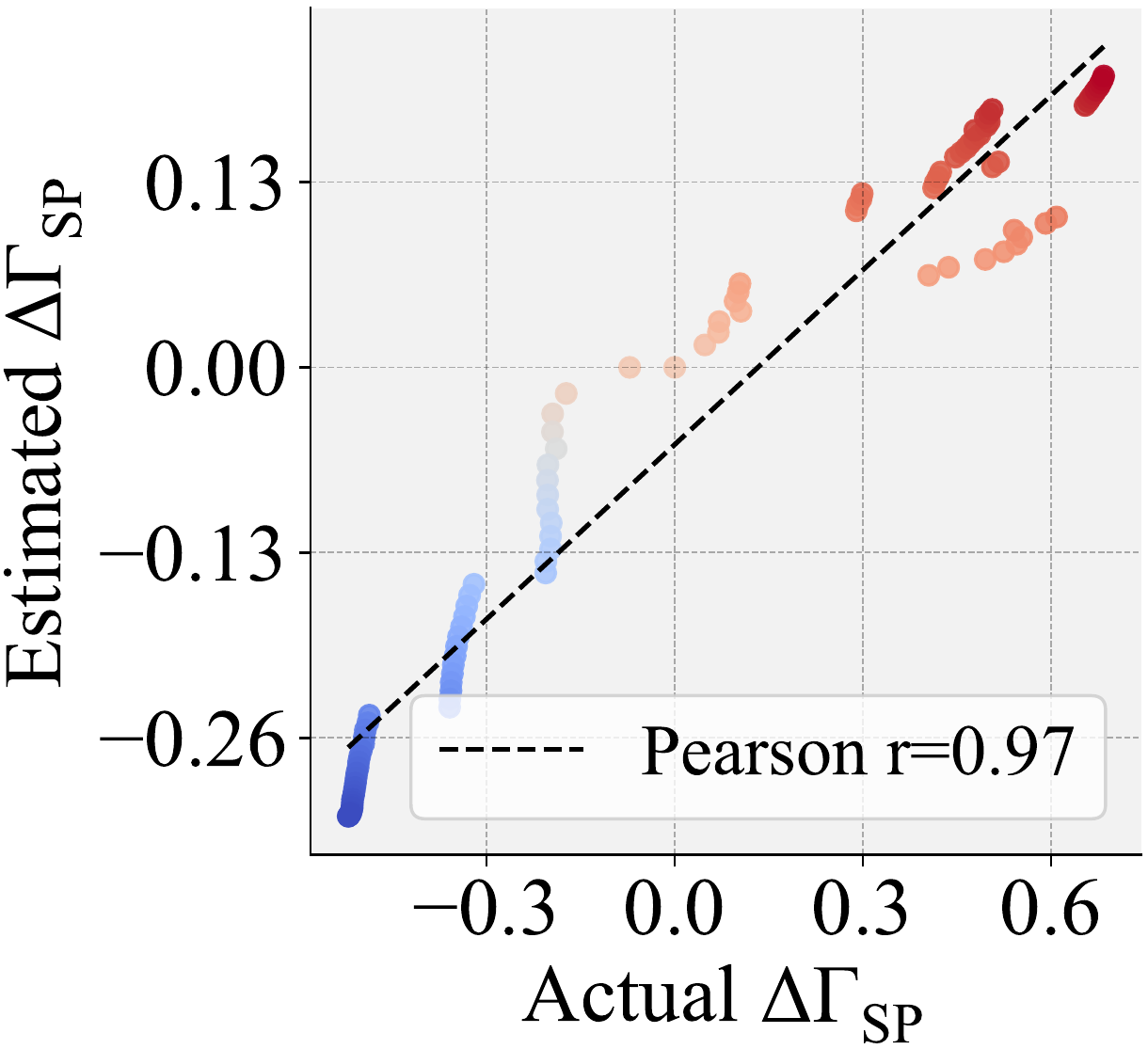}
            \caption[Network2]%
            {{\footnotesize With non-i.i.d. term}} 
            \label{ablation1}
        \end{subfigure}
        \hspace{3.9mm} \begin{subfigure}[t]{0.21\textwidth}
        \small
        \includegraphics[width=1.04\textwidth]{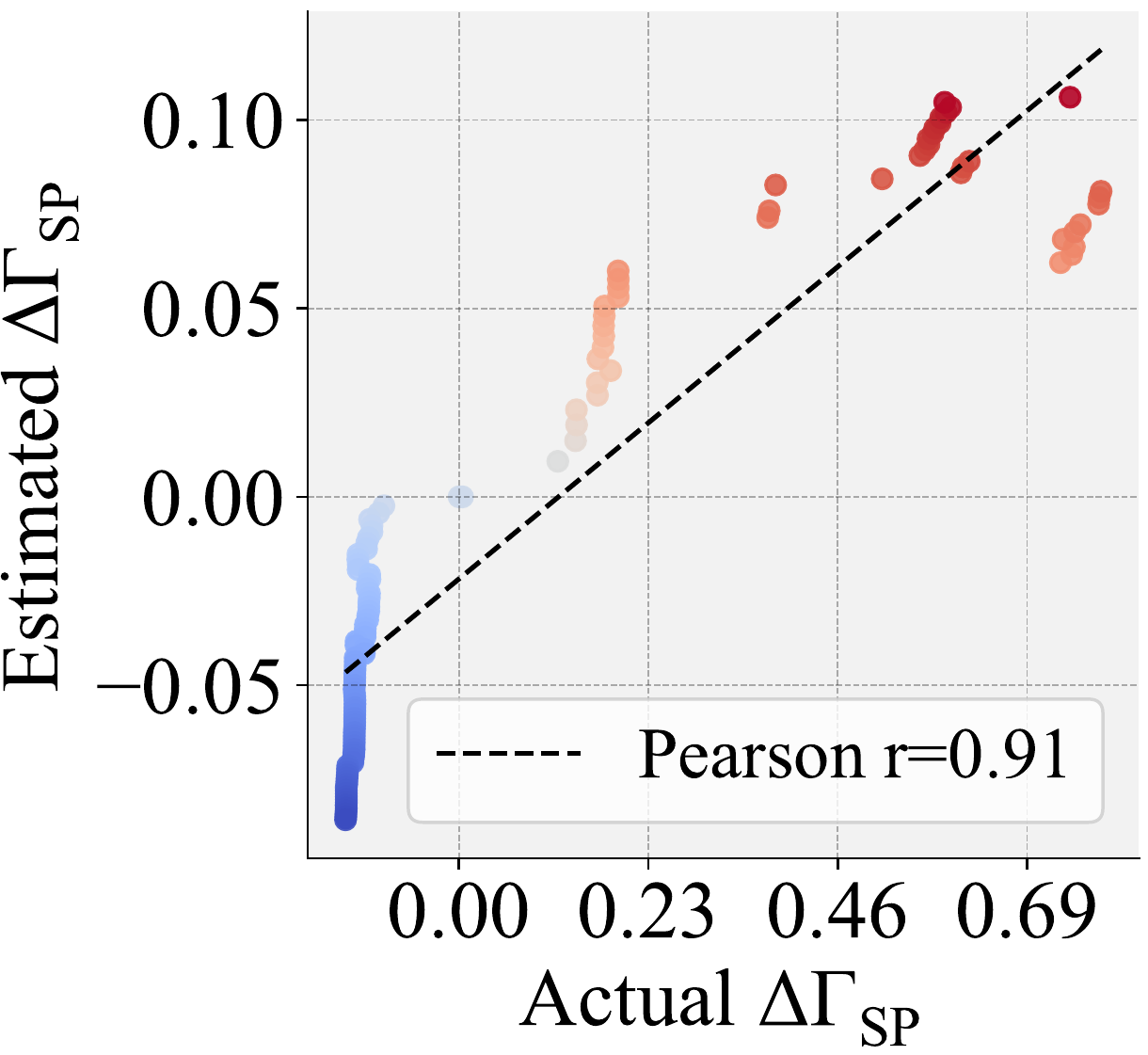}
            \caption[Network2]%
            {{\footnotesize Without non-i.i.d. term}}    
            \label{ablation2}
        \end{subfigure} 
        \hspace{4.2mm}
            \begin{subfigure}[t]{0.21\textwidth}
        \small
        \includegraphics[width=0.9515\textwidth]{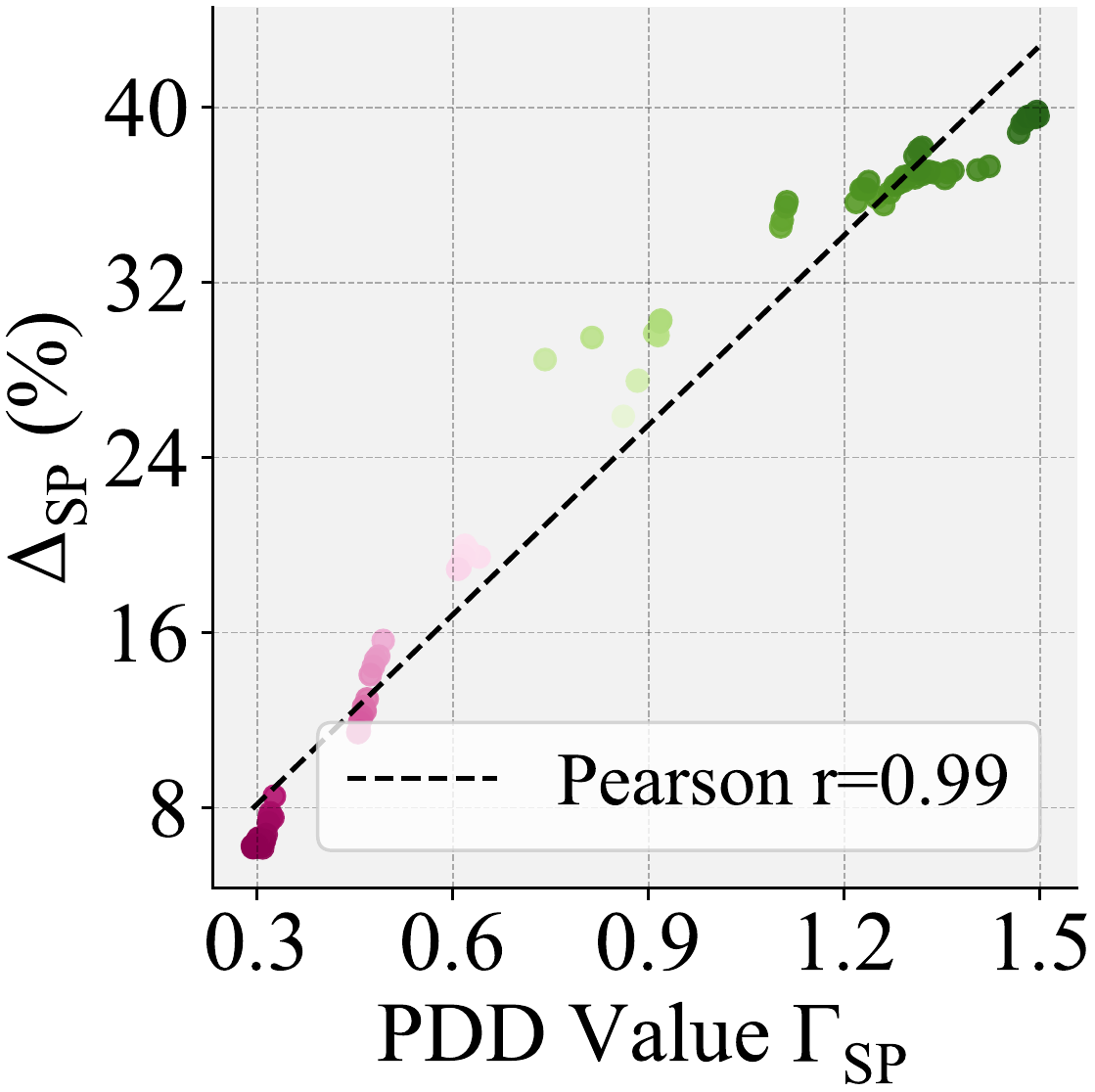}
            \caption[Network2]%
            {{\footnotesize $\Delta_{\text{SP}}$ v.s. $\Gamma_{\text{SP}}$ (Income)}}   
            \label{consis_income}
        \end{subfigure} 
        \hspace{1mm}
        \begin{subfigure}[t]{0.21\textwidth}
        \small
        \includegraphics[width=0.99\textwidth]{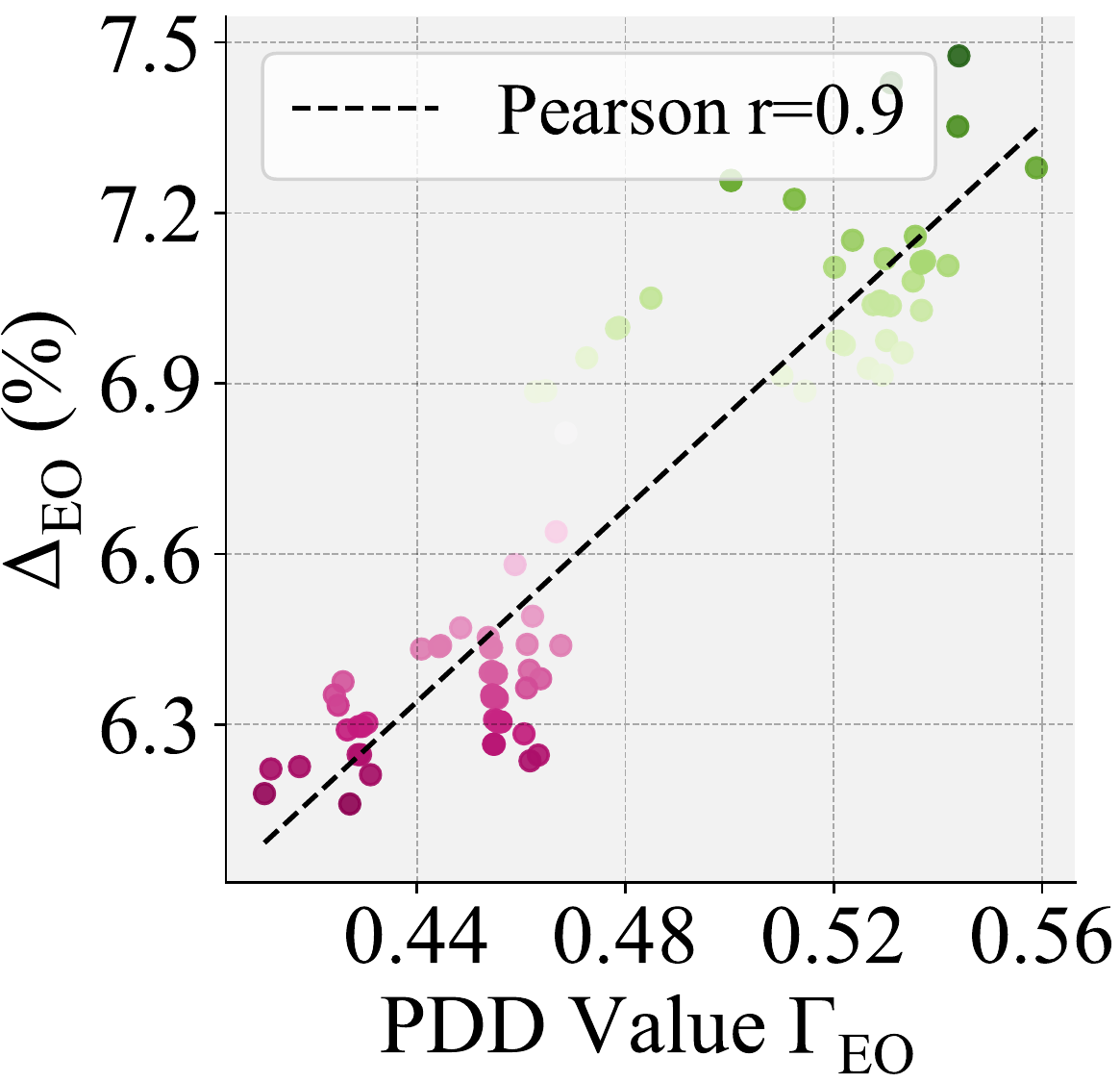}
            \caption[Network2]%
            {{\footnotesize $\Delta_{\text{EO}}$ v.s. $\Gamma_{\text{EO}}$ (Recid.)}} 
            \label{consis_bail}
        \end{subfigure}
            \vspace{-1mm}
    \caption{In (a) and (b), we compare the estimation effectiveness of $\Delta \Gamma_{\text{SP}}$ with and without characterizing non-i.i.d.; in (c) and (d), we present the consistency between $\Gamma$ and traditional fairness metrics ($\Delta_{\text{SP}}$ for statistical parity and $\Delta_{\text{EO}}$ for equal opportunity) under different node deletion budgets.}
    \vspace{-5mm}
    \label{effectiveness-2}
\end{figure*}

Additionally, to understand how the non-i.i.d. characterization benefits the estimation, we also estimate $\Delta \Gamma$ with BIND after the non-i.i.d. characterization being disabled (i.e., setting the $\tilde{L}_{\mathcal{V}_i'}(\mathcal{G}_i, \bm{W})$ term in Eq.~\ref{change} as 0).
We present the estimated $\Delta \Gamma$ v.s. actual $\Delta \Gamma$ on Income dataset with non-i.i.d. characterization being enabled and disabled in Fig.~\ref{ablation1} and~\ref{ablation2}, respectively. 
We observe the correlation decreases between the estimated and actual $\Delta \Gamma$ after the non-i.i.d. characterization is disabled. Such a decrease is also observed on other datasets in terms of both statistical parity and equal opportunity. Such an observation verifies the contribution of non-i.i.d. characterization to the estimation of $\Delta \Gamma$.

Finally, we evaluate how well the values of the proposed PDD matches the values of traditional fairness metrics.
We collect the value pairs of ($\Delta_{\text{SP}}$, $\Gamma_{\text{SP}}$) and ($\Delta_{\text{EO}}$, $\Gamma_{\text{EO}}$) during the GNN re-training process.
The values of $\Delta_{\text{SP}}$ v.s. actual $\Gamma_{\text{SP}}$ are presented in Fig.~\ref{consis_income}, and the values of $\Delta_{\text{EO}}$ v.s. actual $\Gamma_{\text{EO}}$ are shown in Fig.~\ref{consis_bail}. We observe a satisfying match between $\Gamma$ and traditional metrics, which corroborates that PDD is a valid indicator of the fairness level depicted by traditional fairness metrics.

\begin{table*}[]
\vspace{1.5mm}
\caption{Comparison on GNN utility and bias mitigation between BIND and baselines. BIND 1\% and BIND 10\% denote the node deletion budget $k$ being $1\%$ and $10\%$ of the training node set size, respectively. ($\uparrow$) denotes the larger, the better; ($\downarrow$) denotes the opposite. Numerical results are in percentages. Best ones and runner-ups are in \textbf{bold} and \underline{underline}, respectively.}
\label{results}
\centering
\setlength{\extrarowheight}{.11pt}
\setlength\tabcolsep{11pt}
\footnotesize
\begin{tabular}{c|l|cccccc}
\hline
                            &                     & \textbf{Van. GCN} & \textbf{FairGNN} & \textbf{NIFTY} & \textbf{EDITS}  & \textbf{BIND 1\%} & \textbf{BIND 10\%}\\
                            \hline
\multirow{3}{*}{\textbf{Income}}     &($\uparrow$)  \textbf{Acc}         & \underline{74.7 $\pm$ 1.4}    &69.1 $\pm$ 0.6  &70.8 $\pm$ 0.9&68.3 $\pm$ 0.8&\textbf{75.2 $\pm$ 0.0} &71.7 $\pm$ 0.7 \\
                            & ($\downarrow$) \textbf{$\Delta_{\text{SP}}$}  & 25.9 $\pm$ 1.9    &\textbf{12.4 $\pm$ 4.7}  &24.4 $\pm$ 1.6&24.0 $\pm$ 1.9&19.2 $\pm$ 0.6 &\underline{14.7 $\pm$ 1.4}\\
                            & ($\downarrow$) \textbf{$\Delta_{\text{EO}}$}  & 32.3 $\pm$ 0.8    &\textbf{15.6 $\pm$ 6.8}  &26.9 $\pm$ 3.7&24.9 $\pm$ 1.0&26.4 $\pm$ 0.4 &\underline{16.2 $\pm$ 2.0}\\
                            \hline
\multirow{3}{*}{\textbf{Recidivism}} & ($\uparrow$) \textbf{Acc}         & \textbf{89.8 $\pm$ 0.0}    &\underline{89.7 $\pm$ 0.2}  &79.1 $\pm$ 0.9&89.6 $\pm$ 0.1&88.7 $\pm$ 0.0 &88.5 $\pm$ 0.2\\
                            & ($\downarrow$) \textbf{$\Delta_{\text{SP}}$} & 7.47 $\pm$ 0.2    &7.31 $\pm$ 0.5  &\textbf{1.82 $\pm$ 0.8}&\underline{5.02 $\pm$ 0.0}&7.40 $\pm$ 0.0 &6.57 $\pm$ 0.2\\
                            & ($\downarrow$) \textbf{$\Delta_{\text{EO}}$} & 5.23 $\pm$ 0.1    &5.17 $\pm$ 0.0  &\textbf{1.28 $\pm$ 0.5}&\underline{2.89 $\pm$ 0.1}&5.09 $\pm$ 0.1 &4.23 $\pm$ 0.2\\
                            \hline
\multirow{3}{*}{\textbf{Pokec-z}}    & ($\uparrow$) \textbf{Acc}         & 63.2 $\pm$ 0.7    &\underline{64.0 $\pm$ 0.7}  &\textbf{65.3 $\pm$ 0.2}&61.6 $\pm$ 0.9&63.5 $\pm$ 0.4 &62.9 $\pm$ 0.4\\
                            & ($\downarrow$) \textbf{$\Delta_{\text{SP}}$} & 7.32 $\pm$ 2.2    &4.95 $\pm$ 0.8  &2.34 $\pm$ 1.0&\underline{1.29 $\pm$ 0.8}&6.75 $\pm$ 2.3 &\textbf{1.02 $\pm$ 0.9}\\
                            & ($\downarrow$) \textbf{$\Delta_{\text{EO}}$} & 7.60 $\pm$ 2.3    &4.29 $\pm$ 0.7  &\textbf{1.46 $\pm$ 1.3}&\underline{1.62 $\pm$ 1.6}&5.41 $\pm$ 3.4 &2.28 $\pm$ 1.5\\
                            \hline
\multirow{3}{*}{\textbf{Pokec-n}}    & ($\uparrow$) \textbf{Acc}        & 58.5 $\pm$ 0.8    &60.3 $\pm$ 0.5  &\textbf{61.1 $\pm$ 0.3}&56.8 $\pm$ 0.9&\underline{60.6 $\pm$ 0.8} &58.8 $\pm$ 1.8\\
                            & ($\downarrow$) \textbf{$\Delta_{\text{SP}}$} & 6.57 $\pm$ 2.6    &5.30 $\pm$ 1.4  &6.55 $\pm$ 0.7&\underline{2.75 $\pm$ 1.8}&5.85 $\pm$ 2.0 &\textbf{2.45 $\pm$ 0.9} \\
                            & ($\downarrow$) \textbf{$\Delta_{\text{EO}}$} & 2.33 $\pm$ 0.5    &\underline{1.67 $\pm$ 0.2}  &1.83 $\pm$ 0.6&2.24 $\pm$ 1.5&\textbf{1.15 $\pm$ 0.7} &2.22 $\pm$ 1.6 \\
                            \hline
\end{tabular}
\end{table*}

\subsection{Debiasing via Harmful Nodes Deletion}

In this subsection, we demonstrate how BIND could be employed for GNN debiasing.
The basic intuition here is to identify and delete those harmful nodes according to the estimated node influence on model bias, and evaluate whether GNNs can be debiased when they are trained on this new graph.
Specifically, we set $\Gamma = \lambda \Gamma_{\text{SP}} + (1 - \lambda) \Gamma_{\text{EO}}$ and estimate the node influence on $\Gamma$ to consider both statistical parity and equal opportunity.
We then set a budget $k$, and follow the strategy adopted in Section~\ref{effectiveness} to select and delete a set of training nodes with the largest positive influence summation on $\Gamma$ under this budget.
We set $\lambda=0.5$ to assign statistical parity and equal opportunity the same weight, and perform experiments with $k$ being $1\%$ (denoted as BIND 1\%) and $10\%$ (denoted as BIND 10\%) of the total number of training nodes.
We present the results on the four adopted datasets in Table~\ref{results}.
The following observations are made:
(1) compared with other baselines, BIND achieves competitive performance (i.e., lower values) on both $\Delta_{\text{SP}}$ and $\Delta_{\text{EO}}$. 
Hence training GNNs on a new graph after deleting harmful nodes (to fairness) is an effective approach for GNN debiasing;
(2)
there is no obvious performance decrease on the model utility of BIND compared with other baselines. We thus argue that deleting harmful nodes can also lead to a satisfying fairness-utility trade-off.


\section{Related Work}

\noindent \textbf{Graph Neural Networks.}
%
GNNs can be divided into spectral-based and spatial-based ones~\cite{wu2020comprehensive,zhou2020graph}.
Spectral GNNs inherit the insights from Convolutional Neural Networks (CNNs)~\cite{bruna2013spectral}, and it is also followed by many other works~\cite{defferrard2016convolutional,levie2018cayleynets,DBLP:conf/iclr/KipfW17}.
Their goal is to design graph filters to extract task-related information from the input graphs based on Spectral Graph Theory~\cite{chung1997spectral}.
However, spatial GNNs design message-passing mechanisms in the spatial domain to extract information from each node's neighbors~\cite{wu2020comprehensive,zhou2020graph}.
Various aggregation strategies improve their performance on different tasks~\cite{velivckovic2017graph,xu2018powerful,suresh2021breaking,park2020role,he2020lightgcn,schlichtkrull2018modeling}.

\noindent \textbf{Algorithmic Fairness.}
Algorithmic fairness can be defined from different perspectives~\cite{pessach2020algorithmic,mehrabi2021survey,du2020fairness,caton2020fairness,corbett2018measure,mitchell2021algorithmic}, where \textit{Group Fairness} and \textit{Individual Fairness} are two popular notions~\cite{dwork2012fairness}.
Generally, group fairness enforces similar statistics (e.g., positive prediction rate in binary classification tasks) across different demographic subgroups~\cite{dwork2012fairness}. Typically, these demographic subgroups are described by certain sensitive attributes, such as gender, race, and religion.
On the other hand, individual fairness argues for similar outputs for similar individuals~\cite{dwork2012fairness}. 
%
There are many works that enhance algorithmic fairness in different stages of a learning pipeline, including pre-processing~\cite{dong2021edits}, in-processing~\cite{dong2021individual,DBLP:conf/icde/LahotiGW19,dai2021towards}, and post-processing~\cite{kang2020inform}. Particularly, re-weighting training samples to mitigate model bias is a popular fairness-enhancing method during in-processing stage~\cite{wang2022training,han2021balancing,yan2022forml,jiang2020identifying,petrovic2022fair}. %
However, most of these methods only yield a set of weights for training samples to mitigate bias~\cite{yan2022forml,wang2022training}, while to what extent each sample influences the exhibited bias is still unclear.
Different from them, our work aims to understand the influence of each training node on model bias. To the best of our knowledge, this is a first-of-its-kind study.
In addition, most of existing methods based on re-weighting training samples are developed under the IID assumption.
However, in this paper, we also analyze the non-IID characteristic between nodes to better understand how each training node influences model bias.

\noindent \textbf{Interpretation of Deep Learning Models.}
Deep learning models have huge parameter size and high complexity~\cite{buhrmester2021analysis,samek2017explainable,fong2017interpretable,xu2019explainable}. 
To make these models more trustworthy and controllable, many studies have been devoted to improving their transparency~\cite{fong2017interpretable}. Generally, these works are divided into transparency design and post-hoc explanation~\cite{xu2019explainable}. The basic goal of transparency design is to understand the model in terms of model structure~\cite{liu2021learning,zhang2019interpreting} and training algorithms~\cite{plumb2019regularizing}, while post-hoc explanation aims to explain specific prediction results via visualization~\cite{ding2017visualizing} and explanatory examples~\cite{chen2018learning}. 
In the realm of learning on graphs, some existing works aim to interpret GNNs~\cite{ying2019gnnexplainer,luo2020parameterized,yuan2020xgnn}, and they mainly focus on understanding the utility (e.g., node classification accuracy) of GNNs on the test set.
Our work is different from them in two aspects: (1) we focus on interpreting the model bias instead of the utility for GNNs; (2) we aim to understand the model bias via attributing to the training set instead of only focusing on the test set.


\section{Conclusion}
\label{conclusion}

In this paper, we study a novel problem of characterizing how each training node influences the bias exhibited in a trained GNN. We first propose a strategy named Probabilistic Distribution Disparity (PDD), which can be instantiated with different existing fairness notions, to quantify the node influence on the model bias. We then propose a novel framework named BIND to achieve an efficient influence estimation for each training node. We also develop a node deletion strategy to achieve GNN debiasing based on influence estimation. Extensive experiments verify (1) the consistency between the proposed PDD and traditional fairness metrics; (2) the efficiency and effectiveness of the influence estimation algorithm; and (3) the performance of the proposed strategy on GNN debiasing.
We leave interpreting how the unfairness arises in other graph learning tasks as future works.

\section{Acknowledgments}
\label{ack}
This work is supported by the National Science Foundation under grants IIS-2006844, IIS-2144209, IIS-2223768, IIS-2223769, CNS-2154962, and BCS-2228534, the JP Morgan Chase Faculty Research Award, and the Cisco Faculty Research Award.

\linespread{2.0}
\bibliography{ref_renewed}

\linespread{2.0}

\section{Supplementary Discussion}
\label{app_challenges}

\noindent \textbf{Generalized PDD Instantiations.} 
The proposed PDD can be easily generalized to non-binary cases. Here we focus on the generalization to multi-class sensitive attributes and predicted labels.
In multi-class node classification tasks with multi-class sensitive attributes, a widely adopted fairness criterion is to ensure that the ratio of predicted labels under each class is the same across all sensitive subgroups~\cite{DBLP:conf/ijcai/RahmanS0019,DBLP:journals/corr/abs-2104-14210} (i.e., the demographic subgroups that are described by sensitive attributes).
Correspondingly, PDD can be defined as the average Wasserstein-1 distance (of the probabilistic prediction distributions) across all sensitive subgroup pairs. Here the variable of interest is $\hat{\bm{y}}$, and the generalized PDD is formally given as
\begin{align}
    \Gamma_{\text{generalized}} = \frac{2}{|\mathcal{S}| (|\mathcal{S}| - 1)} \sum_{i \neq j \text{and} i, j \in \mathcal{S}} \text{W}_1(P^{(S=i)}_{\hat{\bm{y}}}, P^{(S=j)}_{\hat{\bm{y}}}).
\end{align}
Here $\mathcal{S}$ is the set of all possible values for sensitive attribute $S$; W$_1$($\cdot, \cdot$) takes two probability distributions as input, and outputs the Wasserstein-1 distance between them; $P^{(S=i)}_{\hat{\bm{y}}}$ is the probability distribution of the GNN probabilistic predictions $\hat{\bm{y}}$ with the corresponding $S$ being $i$.



\noindent \textbf{Efficient Estimation of Wasserstein Distance.}
Our first computation challenge lies in the notorious intractability of Wasserstein-1 distance (between two probability distributions).
In additional to its intractability, the derivative of Wasserstein-1 distance between the distributions of two GNN probabilistic prediction sets w.r.t. the GNN parameters is hard to compute.
To tackle both problems, we employ the efficient estimation algorithm proposed in~\cite{cuturi2014fast}, which has been widely adopted to achieve an estimation for both Wasserstein-1 distance and its derivatives w.r.t. the model parameters that are used to generate the data points from the two distributions~\cite{ma2021deconfounding,guo2020learning}.

\noindent \textbf{Efficient Estimation of Hessian Matrix Inverse.}
Our second computation challenge is the high computational cost introduced by the Hessian inverse operation in Eq.~(6). 
To handle such a challenge, we adopt the efficient estimation approach proposed in~\cite{martens2010deep}, which is with linear time complexity~\cite{koh2017understanding,chen2020multi}.
Specifically, this approach estimates the product of the desired Hessian inverse and an arbitrary vector with the same dimension as the columns in the Hessian matrix, which helps to compute the multiplication between $\left(\frac{\partial^{2} L_{\mathcal{V}^{\prime}}(\mathcal{G}, \hat{W})}{\partial W^{2}}\right)^{-1}$ and $\frac{\partial L_{v_{i}}\left(\mathcal{G}_{i}, \hat{\boldsymbol{W}}\right)}{\partial \boldsymbol{W}}$ (or $\frac{\partial \tilde{L}_{\mathcal{V}_{i}^{\prime}}\left(\mathcal{G}_{i}, \hat{W}\right)}{\partial \boldsymbol{W}}$) in Eq.~(6).

\noindent \textbf{Limitations \& Negative Impacts.} The focus of this paper is to interpret how unfairness arises in node classification tasks. However, there could also be unfairness in other graph analytical tasks. We leave this topic for future works. As for the negative impacts of this work, we do not foresee any obvious ones at this moment.

\linespread{1.0}
\begin{table*}[]
\vspace{5mm}
\caption{This table presents the statistics and basic information about the four real-world datasets adopted for experimental evaluation. Sens. represents the semantic meaning of the sensitive attribute.}
\label{datasets}
\centering
\setlength{\extrarowheight}{.11pt}
\setlength\tabcolsep{14pt}
\begin{tabular}{lcccc}
\hline
\textbf{Dataset}          & \textbf{Income}  & \textbf{Recidivism} & \textbf{Pokec-z}   & \textbf{Pokec-n}           \\
\hline
\textbf{\# Nodes}         & 14,821       & 18,876    & 7,659             & 6,185            \\
\textbf{\# Edges}           & 100,483       & 321,308       & 29,476             & 21,844             \\
\textbf{\# Attributes}     & 14       & 18       & 59             & 59            \\
\textbf{Avg. degree}        & 13.6        & 34.0    & 7.70             & 7.06             \\
\textbf{Sens.}            & Race  & Race  & Region             & Region     \\
\textbf{Label}         & Income level  & Bail decision   & Working field             & Working field    \\
\hline
\end{tabular}
\vspace{-1.0em}
\end{table*}
\linespread{1.7}

\section{Implementation Details \& Supplementary Experiments}
\label{app_experiments}

\subsection{Implementation Details}

\noindent \textbf{Real-world Dataset Description.}
There are four real-world datasets adopted in our experiments in total, namely \textit{Income}, \textit{Recidivism}, \textit{Pokec-z}, and \textit{Pokec-n}. For each dataset, there are two classes of ground truth labels (i.e., 0 and 1) for the node classification task. We randomly select 25\% nodes as the validation set and 25\% nodes as the test set, both of which include nodes corresponding to a balanced ratio of ground truth labels.
For the training set, we randomly select either 50\% nodes or 500 nodes in each class of ground truth labels, depending on which is a smaller number.
Such a splitting strategy is also followed by many other works~\cite{agarwal2021towards,dong2021edits}.
We present the statistics of the adopted datasets in Table 1.
A detailed description of these datasets is as follows.

\begin{itemize}  
    \item \textbf{Income.} \textit{Income} is collected from \textit{Adult Data Set}~\cite{Dua:2019}. Each individual is represented by a node. To establish edges between individuals, we first compute the Euclidean distance between every pair of individuals. For the $i$-th individual, we denote its largest Euclidean distance with other individuals as $d_{i}^{(max)}$. Then, we establish connections (i.e., edges) between the $i$-th and $j$-th individuals if their Euclidean distance in the attribute space is larger than 0.7 (edge building threshold) times min($d_{i}^{(max)}$, $d_{j}^{(max)}$). Here function min($\cdot$, $\cdot$) returns the smaller value of the two input scalars. Such a criterion of establishing connections between individuals is a widely adopted strategy in graph-based algorithmic fairness related works~\cite{agarwal2021towards}. The sensitive attribute is race for this dataset, and the task is to classify whether the salary of a person is over \$50K per year or not. 
    \item \textbf{Recidivism.} \textit{Recidivism} is collected from~\cite{jordan2015effect}. A node represents a defendant released on bail, and defendants are connected according to a similar strategy introduced in dataset \textit{Income} with an edge building threshold of 0.6. The sensitive attribute is race, and the task is to classify whether a defendant is on bail or not.
    \item \textbf{Pokec-z \& Pokec-n.} \textit{Pokec-z} and \textit{Pokec-n} are collected from \textit{Pokec}, which is a popular social network in Slovakia~\cite{takac2012data}. In both datasets, each user is a node, and each edge stands for the friendship relation between two users. Different from the two datasets above, the edges in these datasets are directly collected from the original data source. Here the locating region of users is the sensitive attribute, and the task is to classify the working field of users. 
\end{itemize}

\linespread{1.0}
\begin{figure}[t!]
     \centering
     \includegraphics[width=.4\textwidth]{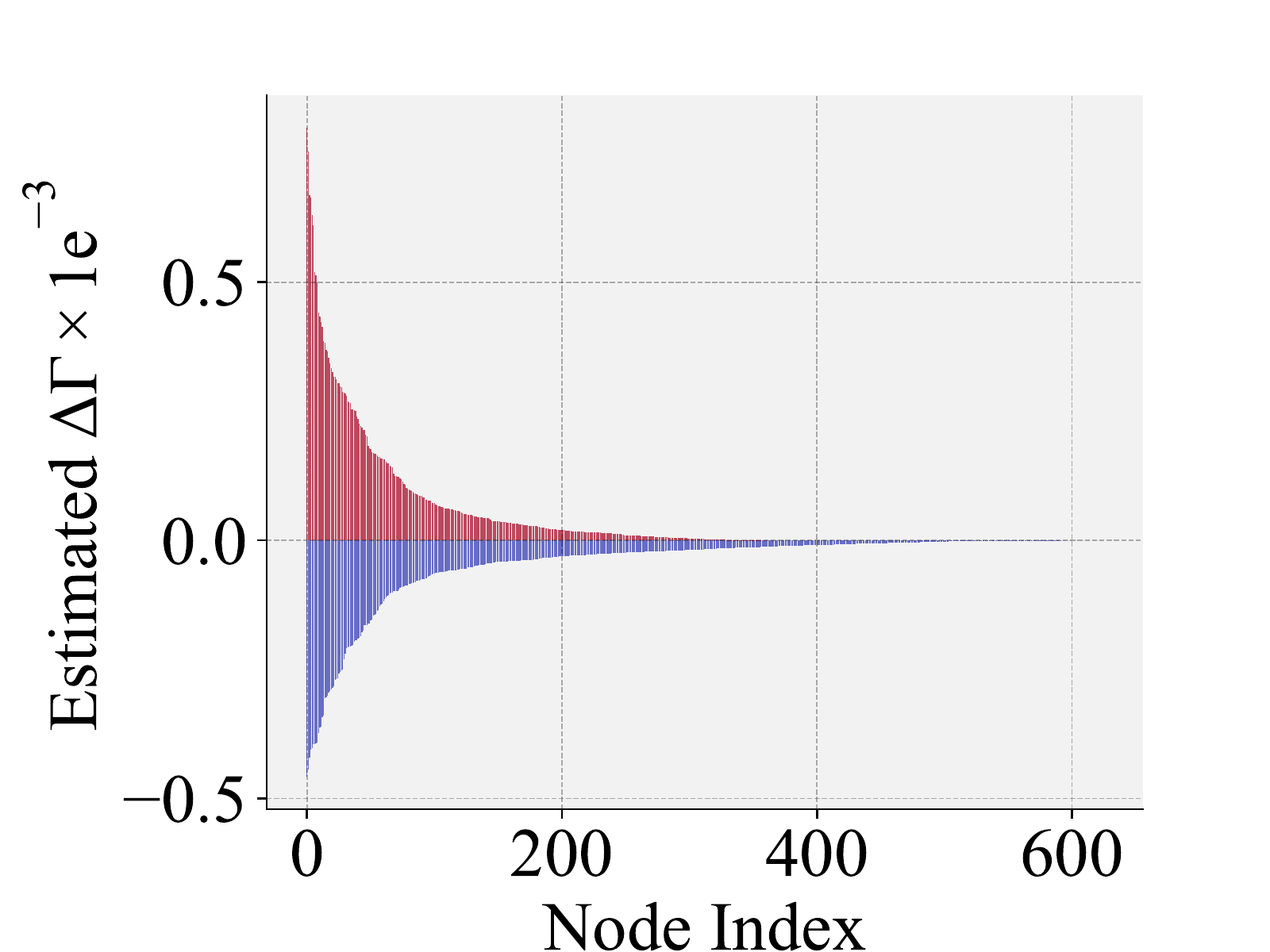}
    \caption{The estimated $\Delta \Gamma_{\text{SP}}$ values follow a long-tail distribution. Here, the color red and blue represent those helpful and harmful nodes, respectively.}
    \label{longtail}
    \vspace{-4mm}
\end{figure}

\begin{figure}
\begin{subfigure}[t]{0.23\textwidth}
\small
\includegraphics[width=0.95\textwidth]{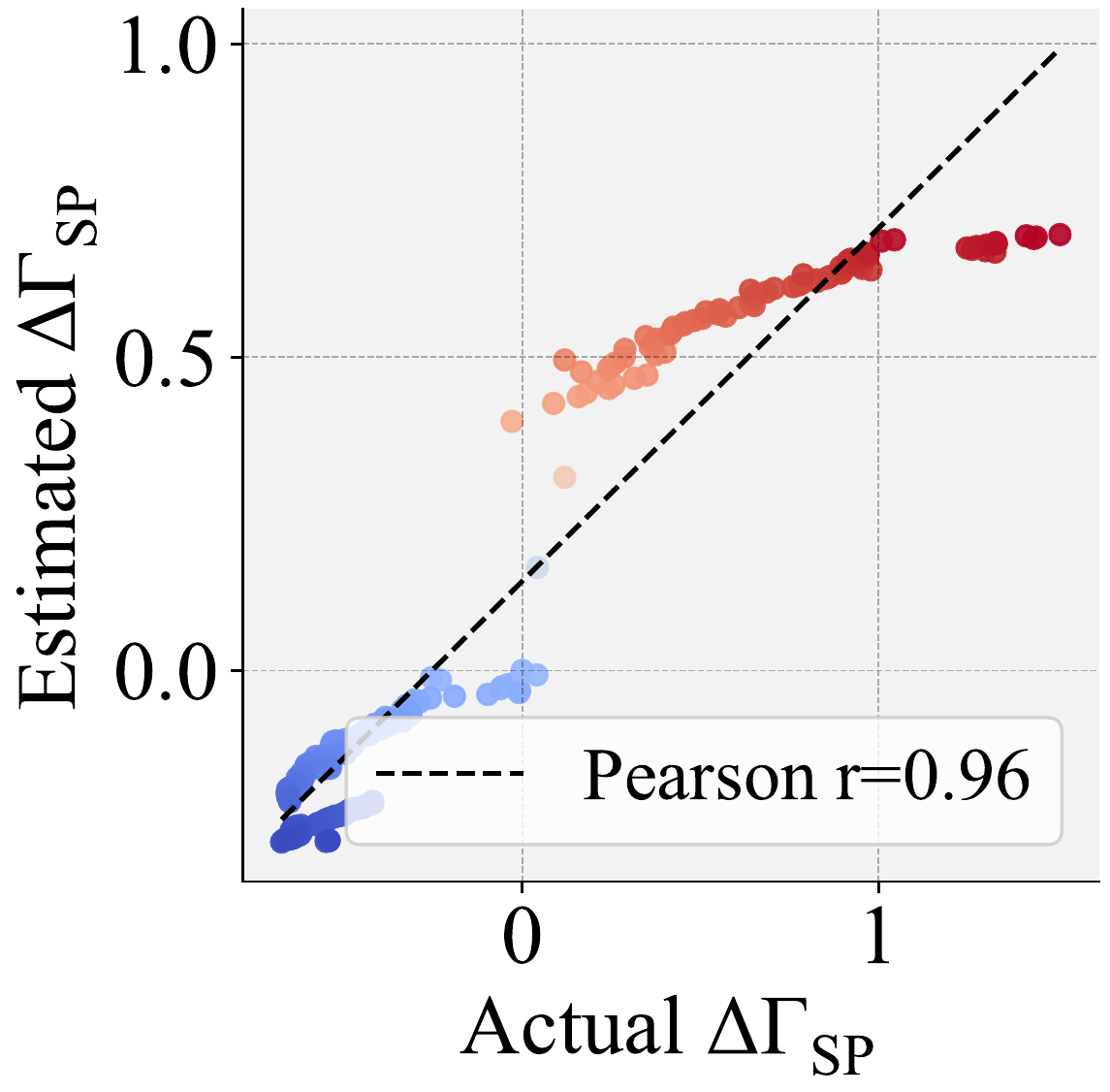}
    \caption[Network2]%
    {{\footnotesize $\Delta \Gamma_{\text{SP}}$ on GIN}}    
    \label{gin}
\end{subfigure} 
        \begin{subfigure}[t]{0.23\textwidth}
\small
\includegraphics[width=0.99\textwidth]{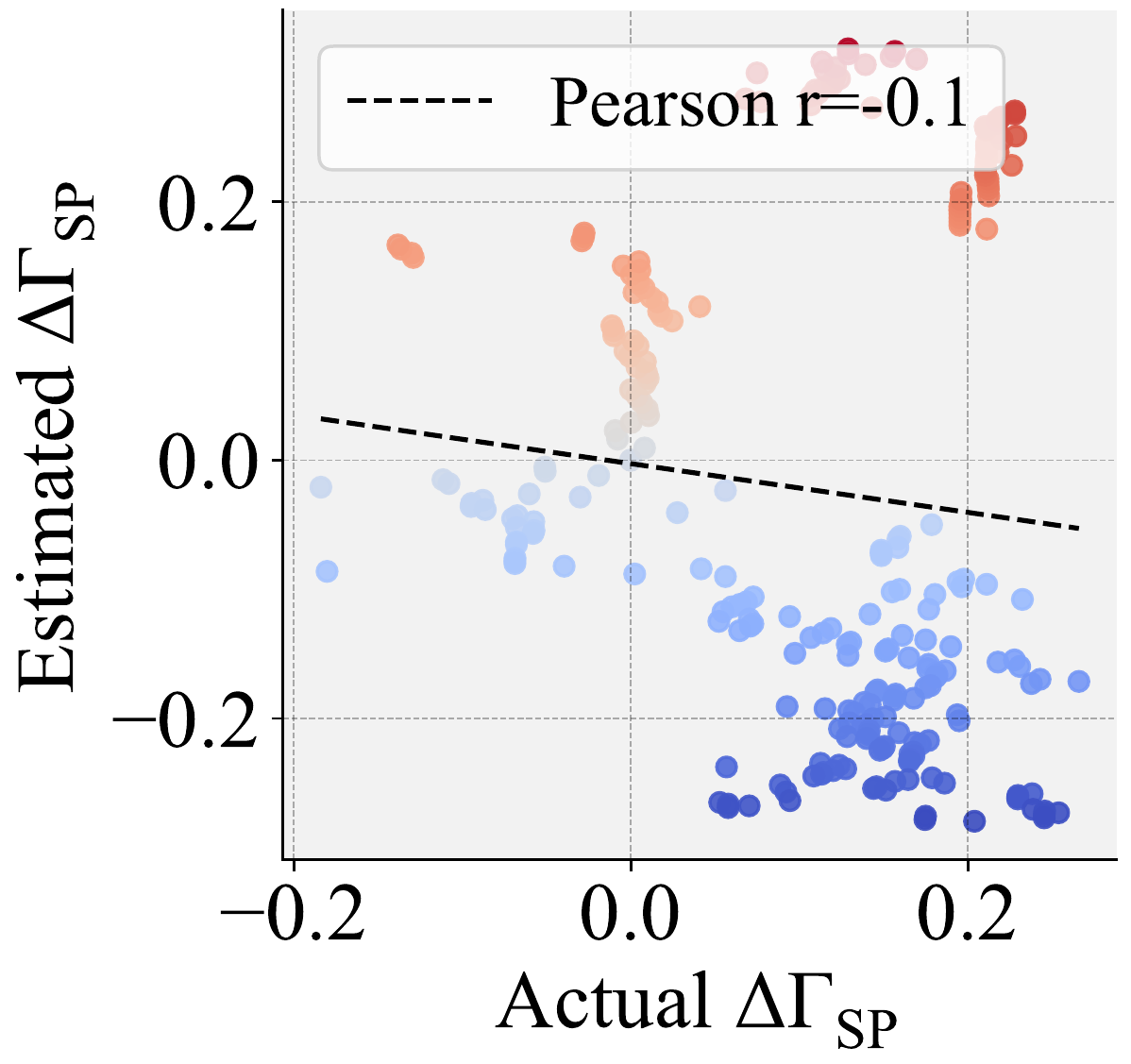}
\vspace{-4mm}
    \caption[Network2]%
    {{\footnotesize Shuffling on GCN}}    
    \label{shuffle}
\end{subfigure}
\caption{In (a), the estimated $\Delta \Gamma_{\text{SP}}$ v.s. the actual $\Delta \Gamma_{\text{SP}}$ based on a GIN model and Income dataset is presented. In (b), we present the estimated $\Delta \Gamma_{\text{SP}}$ v.s. the actual $\Delta \Gamma_{\text{SP}}$ after randomly shuffling the order of estimated $\Delta \Gamma_{\text{SP}}$ values for the training nodes. The correlation value decrease further validates the effectiveness of estimation.}
\label{ablation3}
    \vspace{-4mm}
\end{figure}
\linespread{1.7}

\noindent \textbf{Implementation of GNNs \& BIND.}
We present the implementation of GNNs and BIND as follows.
\begin{itemize}  
    \item \textbf{GNNs.} GNNs are implemented with PyTorch~\cite{paszke2019pytorch}, which is under a BSD-style license. We build all adopted GNNs with one information aggregation layer for simplicity. The information aggregation layers are implemented with PyG (PyTorch Geometric)~\cite{FeyLenssen2019}, which is under an MIT license license. 
    \item \textbf{BIND.} BIND is implemented with PyTorch~\cite{paszke2019pytorch}, Numpy~\cite{harris2020array}, and Scikit-learn~\cite{scikitlearn}, all of which are under a BSD-style license. Code can be found in the supplementary files.
\end{itemize}

\noindent \textbf{Implementation of Baselines.}
For all adopted GNN debiasing baselines, i.e., FairGNN, NIFTY, and EDITS, we adopt their released implementations for a fair comparison. All baselines are implemented with PyTorch~\cite{paszke2019pytorch}. FairGNN and NIFTY are optimized with Adam optimizer~\cite{DBLP:journals/corr/KingmaB14}, while EDITS is optimized with RMSprop~\cite{hinton2012neural} as recommended.

\noindent \textbf{Experimental Settings.}
The experiments in this paper are performed on a workstation with an Intel Exon Silver 4215 (CPU) and a Titan RTX (GPU).
We set seeds as 1, 10, 100 as three runs for the replication of the experiments.
For all GNNs to be interpreted, we train the GNN models for 1000 epochs with the learning rate being 1e-3. The number of the GNN hidden dimension is set as 16 for all experiments, and the dropout rate is set as 0.5. All GNNs are optimized with Adam optimizer~\cite{DBLP:journals/corr/KingmaB14}.
The iteration number for Hessian matrix inverse estimation is chosen from \{10, 50, 100, 500, 1000, 5000\} for the best performance.

\noindent \textbf{Packages Required for Implementations.}
We list main packages and their corresponding versions adopted in our implementations as below. 
\begin{itemize}  
\setlength{\itemsep}{0pt}
    \item Python == 3.8.2
    \item torch == 1.7.1 + cu110
    \item cuda == 11.0
    \item torch-cluster == 1.5.9
    \item torch-geometric == 1.7.0
    \item torch-scatter == 2.0.6
    \item torch-sparse == 0.6.9
    \item tensorboard == 2.4.1
    \item scikit-learn == 0.23.1
    \item numpy == 1.19.5
    \item scipy==1.4.1
    \item networkx == 2.4
\end{itemize}

\linespread{1.0}
\begin{figure*}[!t]
\centering
\vspace{5mm}
        \begin{subfigure}[t]{0.245\textwidth}
        \small
        \includegraphics[width=1.0\textwidth]{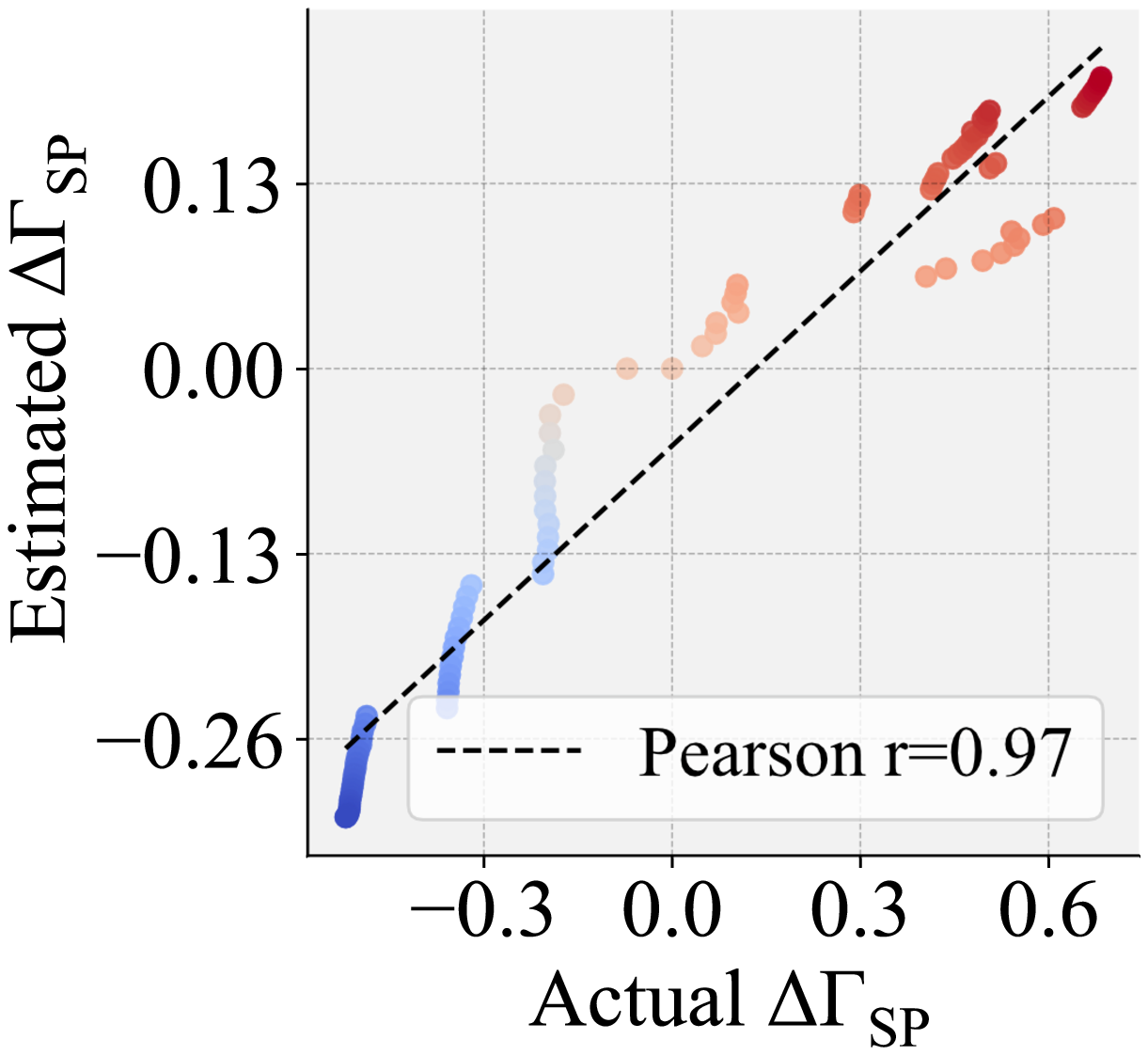}
            \caption[Network2]%
            {{\footnotesize $\Delta \Gamma_{\text{SP}}$ on Income}} 
            \label{sp_income}
        \end{subfigure}
        \centering
        \begin{subfigure}[t]{0.245\textwidth}
        \small
        \includegraphics[width=1.0\textwidth]{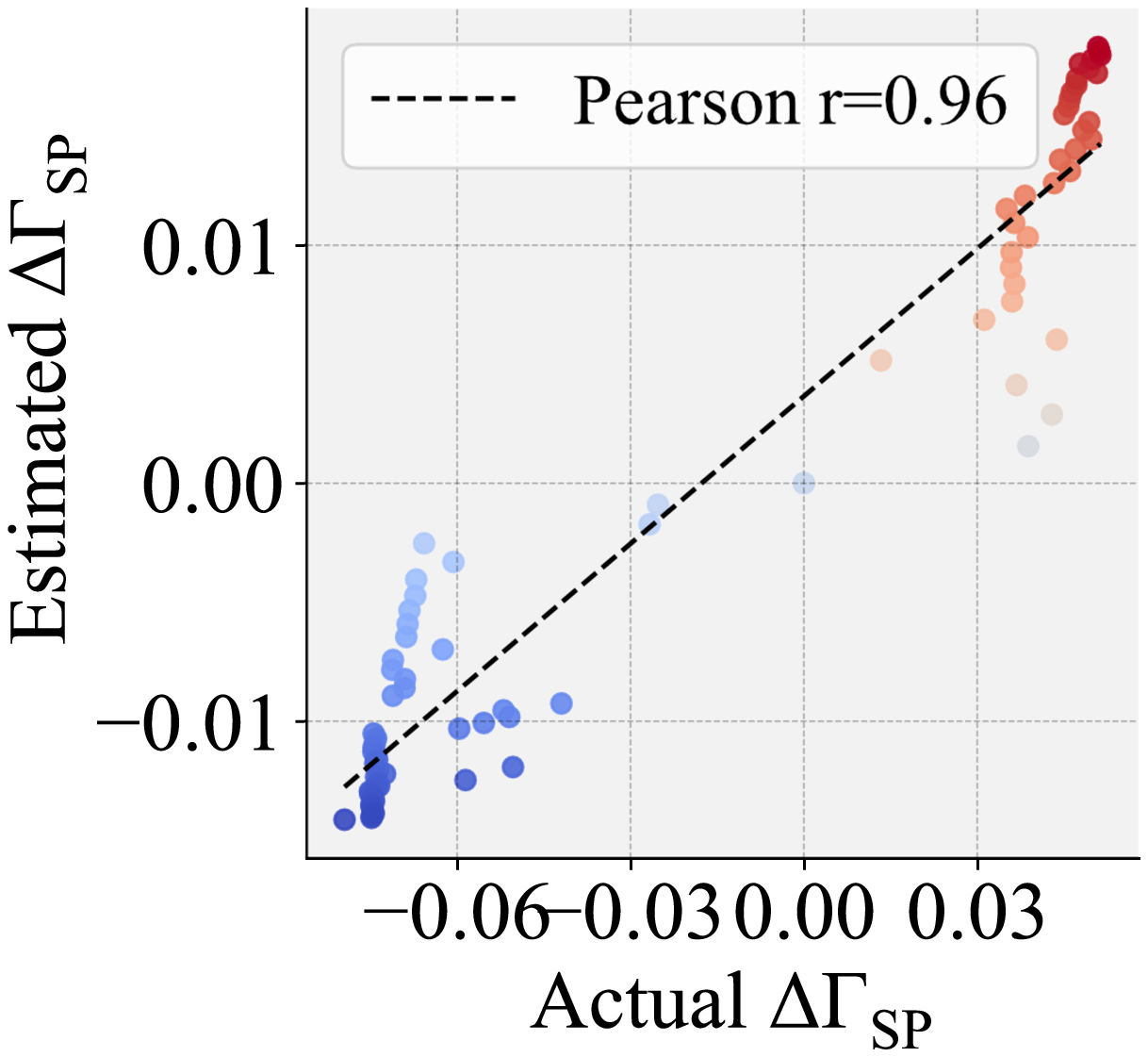}
            \caption[Network2]%
            {{\footnotesize $\Delta \Gamma_{\text{SP}}$ on Recidivism}}   
                        \label{sp_bail}
        \end{subfigure} 
        \begin{subfigure}[t]{0.245\textwidth}
        \small
        \includegraphics[width=1.0\textwidth]{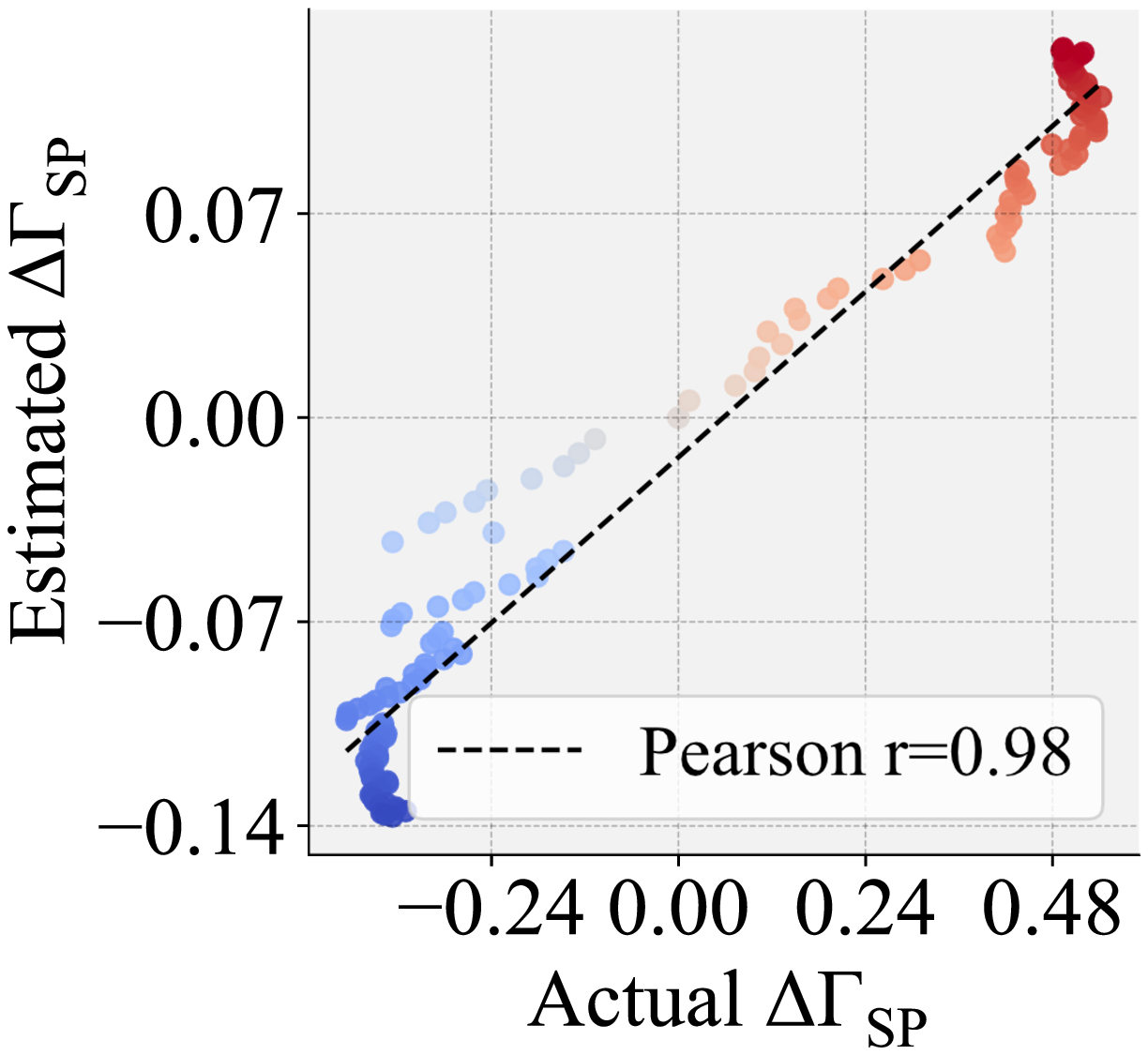}
            \caption[Network2]%
            {{\footnotesize $\Delta \Gamma_{\text{SP}}$ on Pokec-z}}    
                        \label{sp_z}
        \end{subfigure}
        \begin{subfigure}[t]{0.245\textwidth}
        \small
        \includegraphics[width=1.0\textwidth]{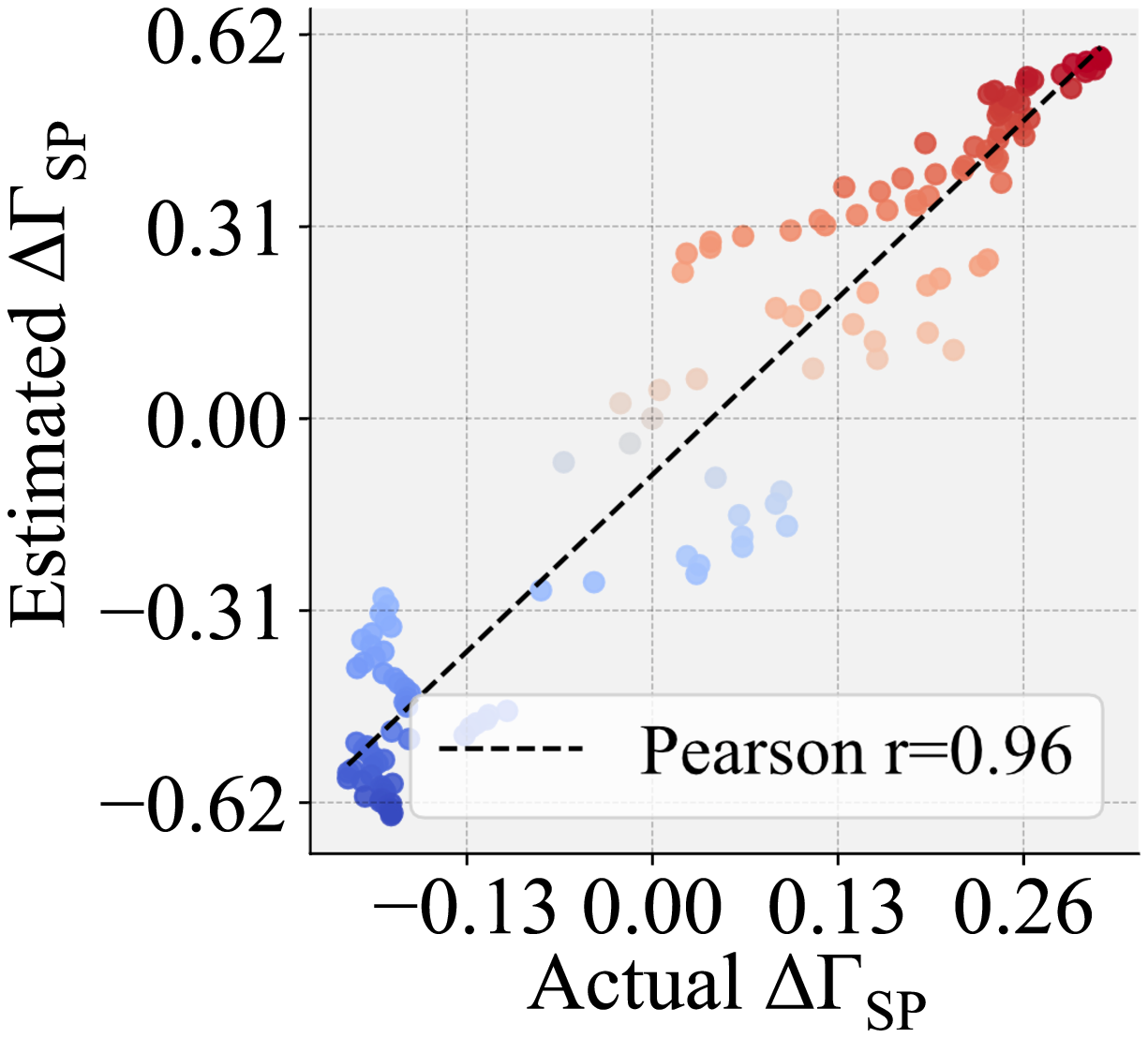}
            \caption[Network2]%
            {{\footnotesize $\Delta \Gamma_{\text{SP}}$ on Pokec-n}}    
                        \label{sp_n}
        \end{subfigure}  \\
        
            \begin{subfigure}[t]{0.245\textwidth}
        \small
        \includegraphics[width=1.0\textwidth]{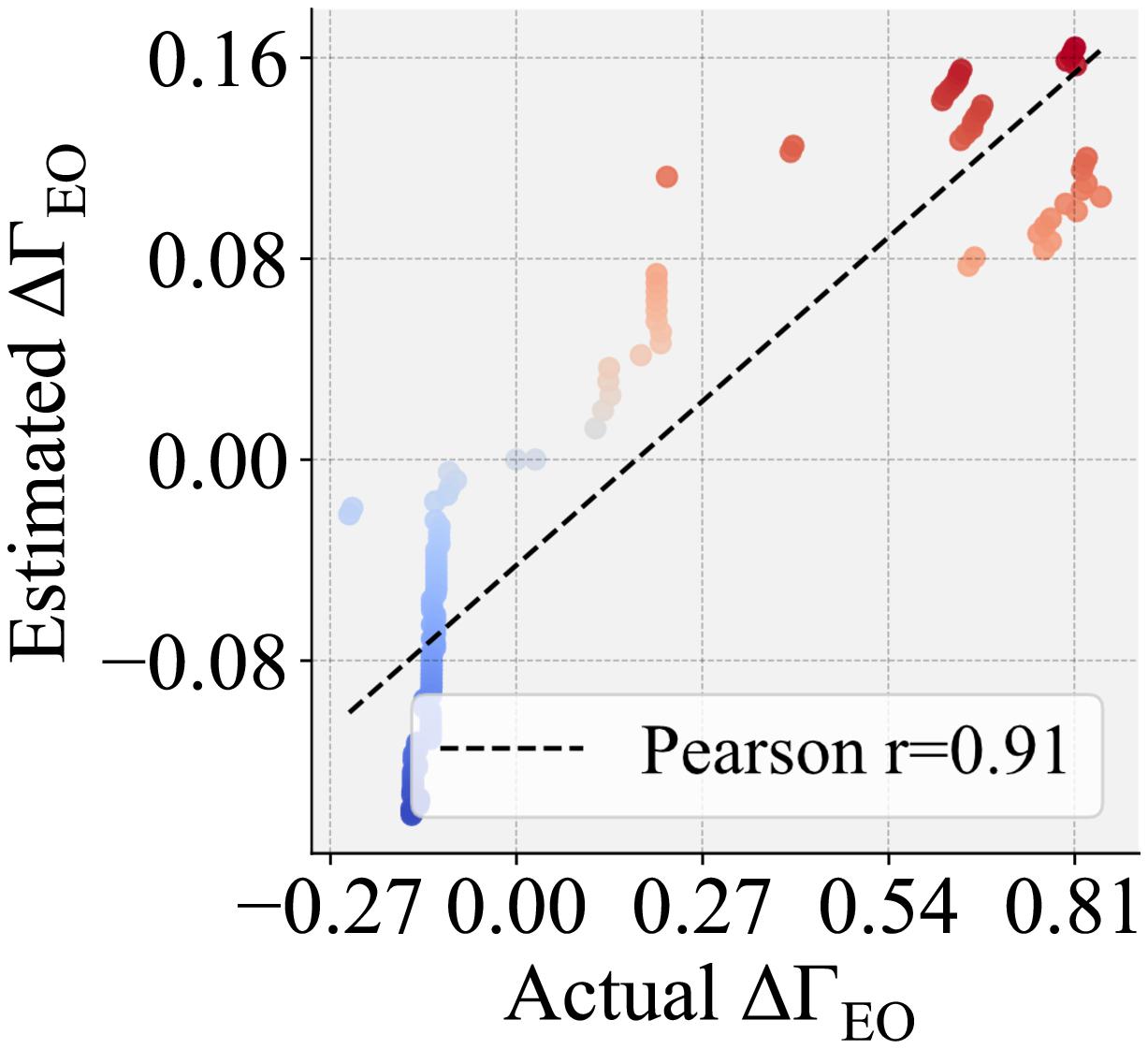}
            \caption[Network2]%
            {{\footnotesize $\Delta \Gamma_{\text{EO}}$ on Income}}    
                        \label{eo_income}
        \end{subfigure}
        \centering
        \begin{subfigure}[t]{0.245\textwidth}
        \small
        \includegraphics[width=1.0\textwidth]{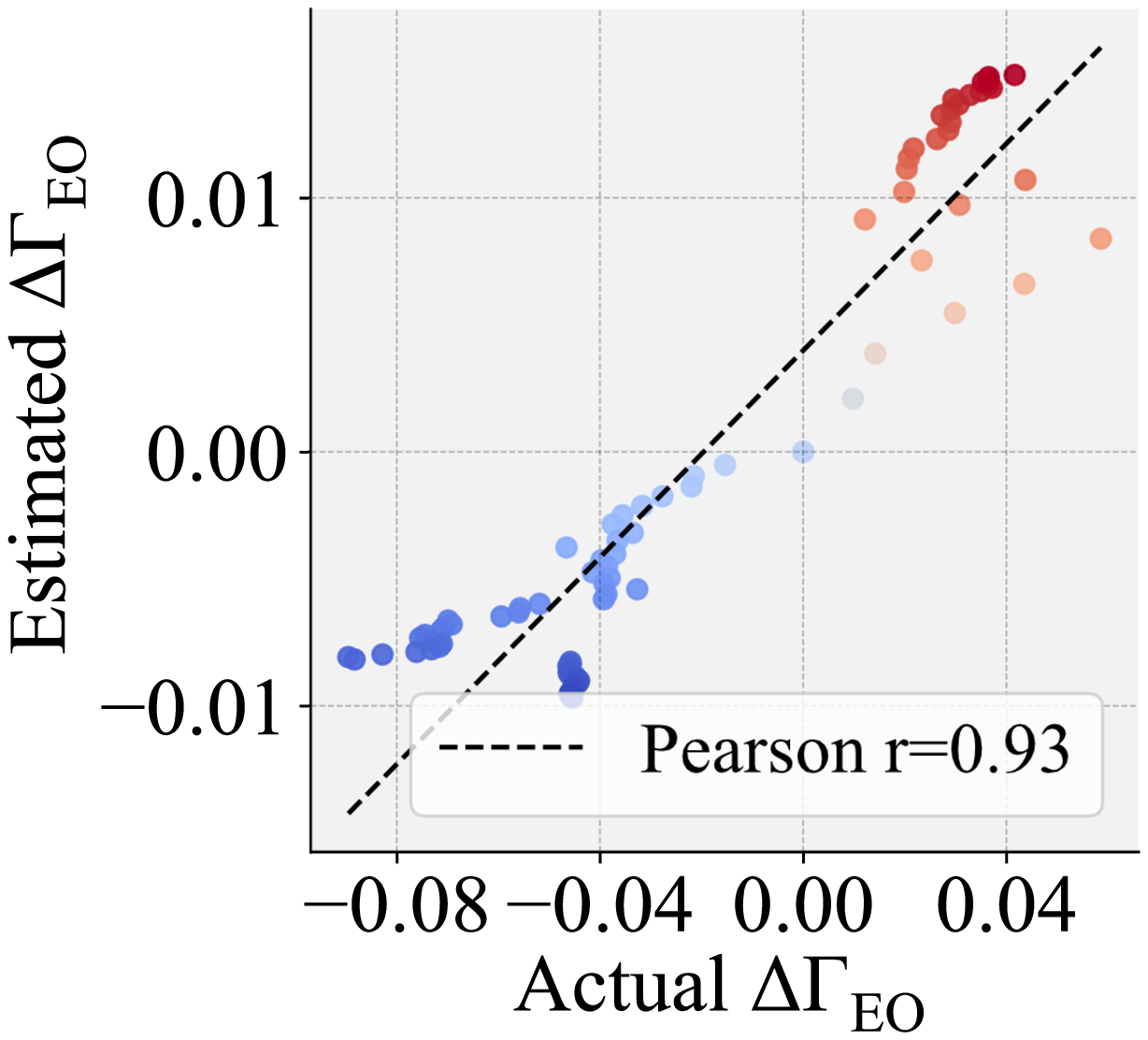}
            \caption[Network2]%
            {{\footnotesize $\Delta \Gamma_{\text{EO}}$ on Recidivism}}    
            \label{eo_bail}
        \end{subfigure} 
        \begin{subfigure}[t]{0.245\textwidth}
        \small
        \includegraphics[width=1.0\textwidth]{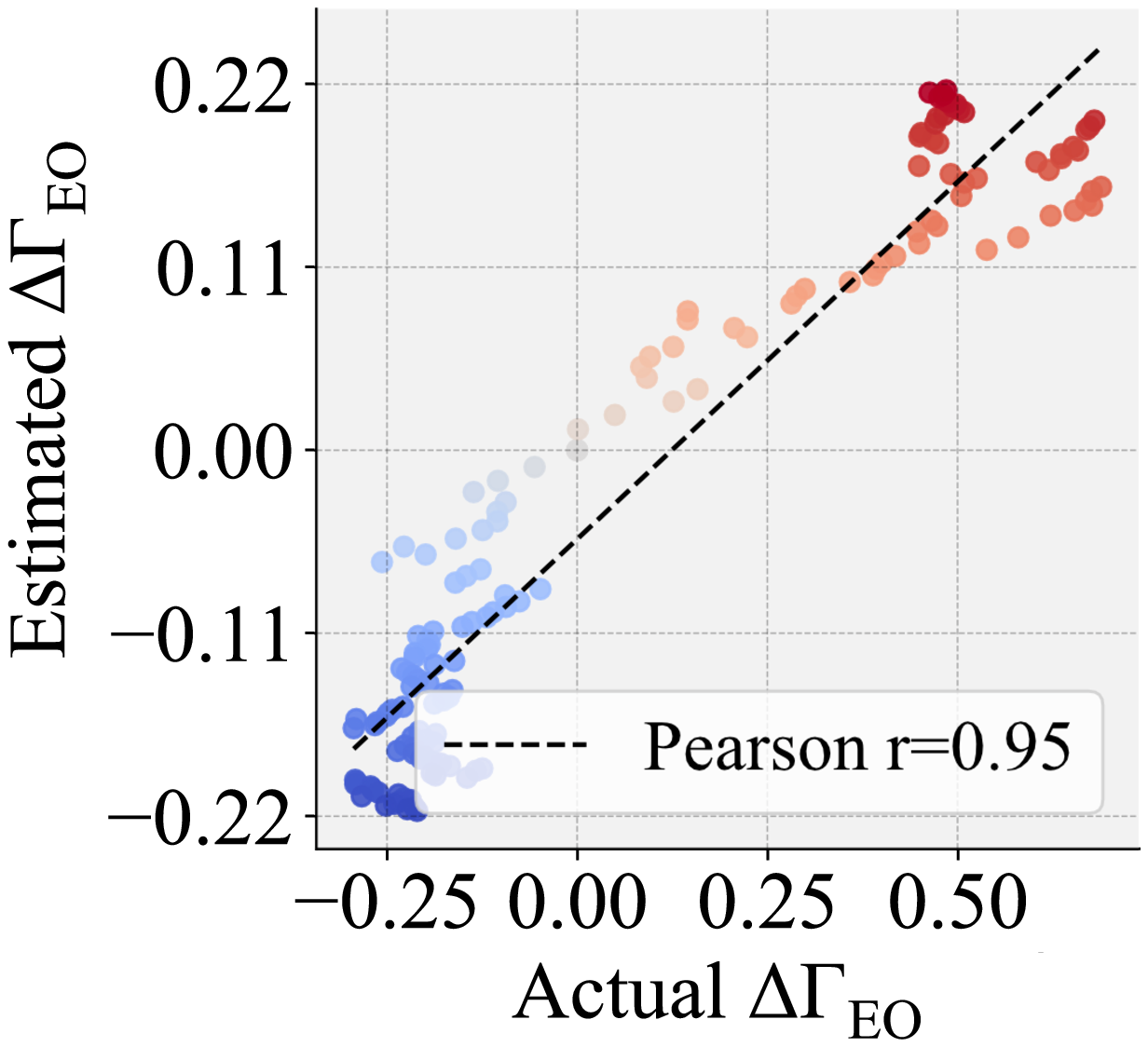}
            \caption[Network2]%
            {{\footnotesize $\Delta \Gamma_{\text{EO}}$ on Pokec-z}}    
                        \label{eo_z}
        \end{subfigure}
        \begin{subfigure}[t]{0.245\textwidth}
        \small
        \includegraphics[width=1.0\textwidth]{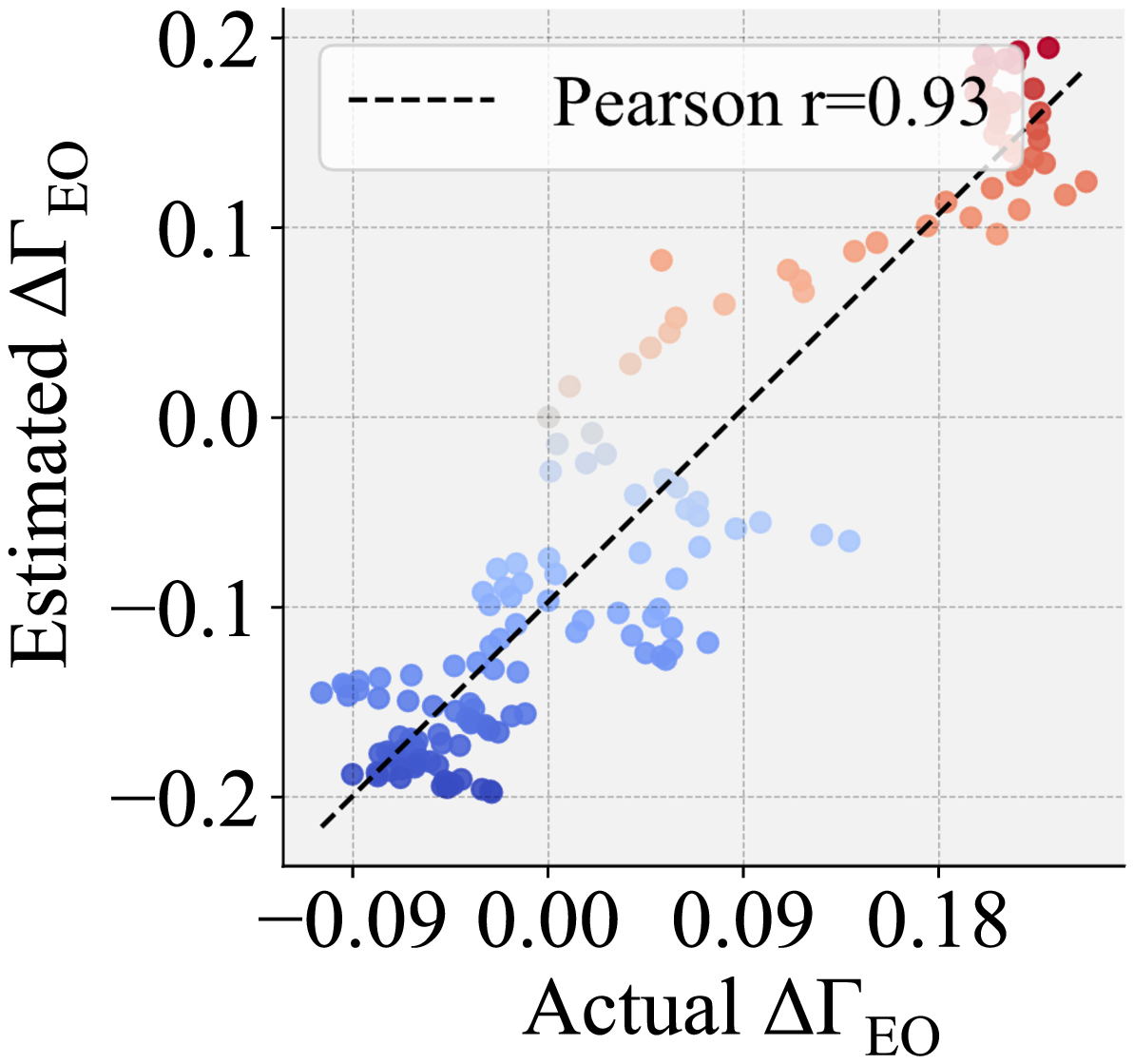}
            \caption[Network2]%
            {{\footnotesize $\Delta \Gamma_{\text{EO}}$ on Pokec-n}}    
                        \label{eo_n}
        \end{subfigure}
    \caption{Estimated $\Delta \Gamma$ v.s. actual $\Delta \Gamma$ on four real-world datasets are presented for effectiveness analysis of $\Delta \Gamma$ estimation based on GCN. Helpful data points (marked in red) are with positive estimated $\Delta \Gamma$ values, while harmful ones (marked in blue) are with negative estimated $\Delta \Gamma$ values.}
    \label{effectiveness-sup2}
\end{figure*}
\linespread{1.7}

\subsection{Supplementary Experiments}

\noindent \textbf{Distribution of Estimated $\Delta \Gamma$.}
We present the estimated $\Delta \Gamma_{\text{SP}}$ on Recidivism dataset in Fig.~\ref{longtail}. Generally, the values of the estimated node influence on bias follow a long-tail distribution, i.e., only a small amount of nodes have large positive/negative influence values on the model bias. Similar phenomena can also be observed on other datasets.


\noindent \textbf{Effectiveness of $\Delta \Gamma$ Estimation.}
We follow the same strategy introduced in the Effectiveness of Node Influence Estimation section to obtain the estimated $\Delta \Gamma_{\text{SP}}$ values and their corresponding actual $\Delta \Gamma_{\text{SP}}$ values.
We present the estimated $\Delta \Gamma_{\text{SP}}$ v.s. the actual $\Delta \Gamma_{\text{SP}}$ on the four datasets in Fig.~\ref{sp_income}, \ref{sp_bail}, \ref{sp_z}, and \ref{sp_n}, respectively; we also present the estimated $\Delta \Gamma_{\text{EO}}$ v.s. the actual $\Delta \Gamma_{\text{EO}}$ on the four datasets in Fig.~\ref{eo_income}, \ref{eo_bail}, \ref{eo_z}, and \ref{eo_n}, respectively.
Experiments are carried out based on a random seed of 42. We draw the conclusion that on both $\Delta \Gamma_{\text{SP}}$ and $\Delta \Gamma_{\text{EO}}$, our estimation results show a satisfying match with the actual values across the four adopted datasets. This validates the effectiveness of $\Delta \Gamma$ estimation.

\noindent \textbf{Generalization of $\Delta \Gamma$ Estimation to Different GNNs.}
We then test the generalization ability of our proposed estimation algorithm to different GNNs. Specifically, we present the estimated $\Delta \Gamma_{\text{SP}}$ v.s. the actual $\Delta \Gamma_{\text{SP}}$ based on a trained GIN model~\cite{DBLP:conf/iclr/XuHLJ19} and Income dataset in Fig.~\ref{gin}. We have the observation that the estimated $\Delta \Gamma_{\text{SP}}$ also shows a satisfying match with the actual values based on the GIN model, which validates the generalization ability of the proposed estimation algorithm to GNNs other than GCN.

\noindent \textbf{Shuffled Node Influence v.s. Actual PDD Difference.}
To further validate the effectiveness of the proposed estimation algorithm, we first perform node influence on bias estimation based on a trained GCN model and Income dataset. Then, we randomly shuffle the estimated influence values (i.e., estimated $\Delta \Gamma_{\text{SP}}$) for the training node set. This operation leads to a mismatch between training node indices and estimated $\Delta \Gamma_{\text{SP}}$. We follow the same strategy introduced in the Effectiveness of Node Influence Estimation section to obtain the estimated $\Delta \Gamma_{\text{SP}}$ values and their corresponding actual $\Delta \Gamma_{\text{SP}}$ values. We present the estimated $\Delta \Gamma_{\text{SP}}$ v.s. actual $\Delta \Gamma_{\text{SP}}$ in Fig.~\ref{shuffle}.
A decrease in Pearson correlation value is observed compared with those presented in Fig.~\ref{effectiveness-sup2}.
This observation further corroborates the effectiveness of the proposed estimation algorithm.

\linespread{1.0}

\section{Proofs}
\label{app_proofs}


\textbf{Theorem 1.}
\textit{According to the optimization objective of $\bm{\hat{W}}_{\epsilon, v_{i}}$ in Eq.~(5), we have
\begin{align}
\left.  \frac{d \bm{\hat{W}}_{\epsilon, v_{i}}}{d \epsilon}\right|_{\epsilon=0}=  &\left(\frac{\partial^{2} L_{\mathcal{V}'}(\mathcal{G},\hat{\bm{W}})}{\partial\bm{W}^{2}}\right)^{-1} \notag \\ &\cdot \left(\frac{\partial L_{v_{i}}\left(\mathcal{G}_{i}, \hat{\boldsymbol{W}}\right)}{\partial \boldsymbol{W}}+\frac{\partial \tilde{L}_{\mathcal{V}_{i}^{\prime}}\left(\mathcal{G}_{i}, \hat{\boldsymbol{W}}\right)}{\partial \boldsymbol{W}}\right).
%
\end{align}  }
\begin{proof}
Here we denote $\frac{1}{m} \sum_{i=1}^{m} L_{v_{i}}\left(\mathcal{G}_{i}, \boldsymbol{W}\right)$ as $L_{\mathcal{V}'}(\mathcal{G}, \bm{W})$. We also have
\begin{align}
\hat{\boldsymbol{W}}_{\epsilon, v_{i}} &\stackrel{\text { def }}{=} \arg \min _{\boldsymbol{W}} L_{\mathcal{V}^{\prime}}(\mathcal{G}, \boldsymbol{W})  \notag \\  &-\epsilon\left(L_{v_{i}}\left(\mathcal{G}_{i}, \boldsymbol{W}\right)+\tilde{L}_{\mathcal{V}_{i}^{\prime}}\left(\mathcal{G}_{i}, \boldsymbol{W}\right)\right).
\end{align}
Consequently, its first-order optimality condition holds, which is given as
\begin{align}
\label{first_order_optimality}
\nabla_{\bm{W}} &L_{\mathcal{V}'}(\mathcal{G}, \hat{\boldsymbol{W}}_{\epsilon, v_{i}}) \notag \\ &- \epsilon \nabla_{\bm{W}} \left( L_{v_i}\left(\mathcal{G}_i,\hat{\boldsymbol{W}}_{\epsilon, v_{i}}\right) + \tilde{L}_{\mathcal{V}_i'}(\mathcal{G}_i,\hat{\boldsymbol{W}}_{\epsilon, v_{i}})\right) = 0.
\end{align}
We define 
\begin{align}
\Psi(\hat{\boldsymbol{W}}_{\epsilon, v_{i}}) &\stackrel{\text { def }}{=} L_{\mathcal{V}'}(\mathcal{G}, \hat{\boldsymbol{W}}_{\epsilon, v_{i}}) \notag \\ & - \epsilon \left( L_{v_i}\left(\mathcal{G}_i,\hat{\boldsymbol{W}}_{\epsilon, v_{i}}\right) + \tilde{L}_{\mathcal{V}_i'}(\mathcal{G}_i,\hat{\boldsymbol{W}}_{\epsilon, v_{i}})\right)
\end{align}
for simplicity. Recall that 
\begin{align}
\hat{\boldsymbol{W}} \stackrel{\text { def }}{=} \arg \min _{\boldsymbol{W}} L_{\mathcal{V}^{\prime}}(\mathcal{G}, \boldsymbol{W}).
\end{align}
When $\epsilon \rightarrow 0$, $\hat{\boldsymbol{W}}_{\epsilon, v_{i}} \rightarrow \hat{\boldsymbol{W}}$. According to Taylor expansion, we have 
\begin{align}
\label{delta_solving}
\nabla_{\bm{W}} \Psi(\hat{\boldsymbol{W}}) + \Delta_{\epsilon, \bm{W}} \nabla_{\bm{W}}^{2} \Psi(\hat{\boldsymbol{W}}) \approx 0
\end{align}
based on Eq.~(\ref{first_order_optimality}), where $\Delta_{\epsilon, \bm{W}} = \hat{\boldsymbol{W}}_{\epsilon, v_{i}} - \hat{\boldsymbol{W}}$. 
We then can solve Eq.~(\ref{delta_solving}) as 
\begin{align}
\Delta_{\epsilon, \bm{W}} \approx  - \left( \nabla_{\bm{W}}^{2} \Psi(\hat{\boldsymbol{W}}) \right)^{-1} \nabla_{\bm{W}} \Psi(\hat{\boldsymbol{W}})
\end{align}
Following the simplification introduced by~\cite{koh2017understanding}, we drop $o$($\epsilon$) terms and let $\nabla_{\bm{W}} L_{\mathcal{V}'}(\mathcal{G}, \hat{\boldsymbol{W}}) = \bm{0}$. We then have
\begin{align}
\Delta_{\epsilon, \bm{W}} \approx  & \epsilon \left(\frac{\partial^{2} L_{\mathcal{V}^{\prime}}(\mathcal{G}, \hat{W})}{\partial \boldsymbol{W}^{2}}\right)^{-1} \notag \\ &\cdot \left(\frac{\partial L_{v_{i}}\left(\mathcal{G}_{i}, \hat{\boldsymbol{W}}\right)}{\partial \boldsymbol{W}}+\frac{\partial \tilde{L}_{\mathcal{V}_{i}^{\prime}}\left(\mathcal{G}_{i}, \hat{\boldsymbol{W}}\right)}{\partial \boldsymbol{W}}\right).
\end{align}
Consequently, we have
\begin{align}
\left. \frac{d \Delta_{\epsilon, \bm{W}}}{d \epsilon}\right|_{\epsilon=0} = & \left(\frac{\partial^{2} L_{\mathcal{V}^{\prime}}(\mathcal{G}, \hat{W})}{\partial \boldsymbol{W}^{2}}\right)^{-1} \notag \\ &\cdot \left(\frac{\partial L_{v_{i}}\left(\mathcal{G}_{i}, \hat{\boldsymbol{W}}\right)}{\partial \boldsymbol{W}}+\frac{\partial \tilde{L}_{\mathcal{V}_{i}^{\prime}}\left(\mathcal{G}_{i}, \hat{\boldsymbol{W}}\right)}{\partial \boldsymbol{W}}\right).
\end{align}
Therefore, we have 
\begin{align}
\left. \frac{d \Delta_{\epsilon, \bm{W}}}{d \epsilon}\right|_{\epsilon=0}  &= \left.\frac{d \hat{\boldsymbol{W}}_{\epsilon, v_{i}}}{d \epsilon}\right|_{\epsilon=0} - 0 \notag \\ & =  \left(\frac{\partial^{2} L_{\mathcal{V}^{\prime}}(\mathcal{G}, \hat{\boldsymbol{W}})}{\partial \boldsymbol{W}^{2}}\right)^{-1}  \notag \\ &\cdot \left(\frac{\partial L_{v_{i}}\left(\mathcal{G}_{i}, \hat{\boldsymbol{W}}\right)}{\partial \boldsymbol{W}}+\frac{\partial \tilde{L}_{\mathcal{V}_{i}^{\prime}}\left(\mathcal{G}_{i}, \hat{\boldsymbol{W}}\right)}{\partial \boldsymbol{W}}\right).
\end{align}
\end{proof}

\linespread{1.0}

\textbf{Corollary 1.}
\textit{Define the derivative of $\Gamma$ w.r.t. $\epsilon$ at $\epsilon=0$ as $I_{\Gamma}(v_i)$.
%
%
According to Theorem 1, we have
\begin{align}
\left. I_{\Gamma}(v_i) \stackrel{\text { def }}{=}  \frac{\partial \Gamma}{\partial \epsilon}  \right|_{\epsilon=0} =  \left(\frac{\partial \Gamma}{\partial \bm{W}} \right)^{\top} \left.  \frac{d \bm{\hat{W}}_{\epsilon, v_{i}}}{d \epsilon}\right|_{\epsilon=0}.
\end{align}}
\begin{proof}
$\Gamma$ is a function of $\bm{W}$, as $\Gamma$ is a function of test node representations and these representations are directly calculated based on $\bm{W}$. Assume $\Gamma$ is differentiable w.r.t. $\bm{W}$ (refer to Appendix A for the validity of this assumption). 
At the same time, we know $\bm{\hat{W}}_{\epsilon, v_{i}}$ is also a function of $\epsilon$, and the derivative of $\bm{\hat{W}}_{\epsilon, v_{i}}$ w.r.t. $\epsilon$ exits according to Theorem 1.
As a consequence, $\Gamma$ is a function of $\epsilon$, and $\left. \frac{\partial \Gamma}{\partial \epsilon}  \right|_{\epsilon=0} =  \left(\frac{\partial \Gamma}{\partial \bm{W}} \right)^{\top} \left.  \frac{d \bm{\hat{W}}_{\epsilon, v_{i}}}{d \epsilon}\right|_{\epsilon=0}$ according to the chain rule.
\end{proof}

\textbf{Theorem 2.}
\textit{Compared with the GNN trained on $\mathcal{G}$, $\Delta \Gamma = \Gamma_{\frac{1}{m},v_i} - \Gamma_{0,v_i}$ is equivalent to the value change in $\Gamma$ when the GNN model is trained on graph $\mathcal{G}_{-i}$.}
\begin{proof}
Recall that 
\begin{align}
\hat{\boldsymbol{W}}_{\epsilon, v_{i}} &\stackrel{\text { def }}{=} \arg \min _{\boldsymbol{W}} L_{\mathcal{V}^{\prime}}(\mathcal{G}, \boldsymbol{W})  \notag \\ & -\epsilon\left(L_{v_{i}}\left(\mathcal{G}_{i}, \boldsymbol{W}\right)+\tilde{L}_{\mathcal{V}_{i}^{\prime}}\left(\mathcal{G}_{i}, \boldsymbol{W}\right)\right).
\end{align}
When $\epsilon = \frac{1}{m}$, we define 
\begin{align}
\Omega(\frac{1}{m}) \stackrel{\text { def }}{=} L_{\mathcal{V}^{\prime}}(\mathcal{G}, \boldsymbol{W})- \frac{1}{m} L_{v_{i}}\left(\mathcal{G}_{i}, \boldsymbol{W}\right) - \frac{1}{m} \tilde{L}_{\mathcal{V}_{i}^{\prime}}\left(\mathcal{G}_{i}, \boldsymbol{W}\right).
\end{align}
Then, we have
\begin{align}
\Omega(\frac{1}{m}) &= L_{\mathcal{V}^{\prime}}(\mathcal{G}, \boldsymbol{W})- \frac{1}{m} L_{v_{i}}\left(\mathcal{G}_{i}, \boldsymbol{W}\right) \notag \\ & \;\;\;\; - \frac{1}{m}  \sum_{v_{j} \in \mathcal{V}_{i}^{\prime} \backslash\left\{v_{i}\right\}} \left( L_{v_{j}}\left(\mathcal{G}_{j}, \boldsymbol{W}\right) -  L_{v_{j}}\left(\mathcal{G}_{j,-i}, \boldsymbol{W}\right) \right) \notag \\
&= \frac{1}{m} \sum_{v_{j} \notin \mathcal{V}_{i}^{\prime}} L_{v_{j}}\left(\mathcal{G}_{j}, \boldsymbol{W}\right) \notag \\
& \;\;\;\; + \frac{1}{m} \sum_{v_{j} \in \mathcal{V}_{i}^{\prime} \backslash\left\{v_{i}\right\}} L_{v_{j}}\left(\mathcal{G}_{j,-i}, \boldsymbol{W}\right),
\end{align}
where $\mathcal{G}_{j,-i}$ is the computation graph of $v_j$ with $v_i$ being removed. As a consequence, $\Omega(\frac{1}{m})$ is the loss function based on $\mathcal{G}_{-i}$, i.e., $\mathcal{G}$ with node $v_i$ being removed.
Correspondingly, $\Delta \Gamma = \Gamma_{\frac{1}{m},v_i} - \Gamma_{0,v_i}$ is equivalent to the value change in $\Gamma$ when the GNN model is trained on graph $\mathcal{G}_{-i}$.
\end{proof}

\textbf{Proposition 1.} 
\textit{Denote the learned representations of node $v_j (v_j \in \mathcal{V} \backslash \mathcal{V}')$ based on $\mathcal{G}$ and $\mathcal{G}_{-i}$ as $\bm{z}_j$ and $\bm{z}_j^{\star}$, respectively.
Define $h^{(j,i)}$ and $q^{(j,i)}$ as the distance from $v_j$ to $v_i$ and the number of all possible paths from $v_j$ to $v_i$, respectively.
Define the set of geometric mean node degrees of $q^{(j,i)}$ paths as $\mathcal{D} = \{d_{1}^{(j,i)}, ..., d_{q^{(j,i)}}^{(j,i)}\}$.
Define $d_{min}^{(j,i)}$ as the minimum value of $\mathcal{D}$.
Assume the norms of all node representations are the same.
We then have $\|\bm{z}_j^{\star}-\bm{z}_j\|_2/\|\bm{z}_j\|_2\leq q^{(j,i)}/(d_{min}^{(j,i)})^{h^{(j,i)}}$.}
\begin{proof}
In the following proof, we will follow \cite{huang2020graph} and \cite{xu2018representation} to use GCNs as the exemplar GNN of this proof for simplicity. However, our proof can also be naturally generalized to other GNNs (e.g., GAT~\cite{velivckovic2017graph} and GraphSAGE~\cite{hamilton2017inductive}) by choosing different values for edge weights. Specifically, the $l$-th layer propagation process can be represented as $\mathbf{H}^{(l+1)}=\sigma(\hat{\mathbf{A}}\mathbf{H}^{(l)}\mathbf{W}^{(l})$, where $\mathbf{H}^{(l)}$ and $\mathbf{W}^{(l)}$ denote the node representation and weight parameter matrices, respectively. $\hat{\mathbf{A}}=\mathbf{D}^{-1}\mathbf{A}$ is the adjacency matrix after row normalization. Following \cite{huang2020graph}, \cite{wang2020unifying}, and \cite{xu2018representation}, we set $\sigma$ as an identity function and $\mathbf{W}$ an identity matrix. 
According to the Taylor's theorem, we have
\begin{equation}
    \begin{aligned}
    \bm{z}_j^{\star}-\bm{z}_j\approx\frac{\partial\bm{z}_j}{\partial\bm{z}_i}\bm{z}_i
    \end{aligned}.
\end{equation}
To calculate $\frac{\partial\bm{z}_j}{\partial\bm{z}_i}$, we can expand the $\bm{z}_j$ according to the propagation rule. Then we have
\begin{align}
    \frac{\partial\bm{z}_j}{\partial\bm{z}_i}
    &=\frac{\partial}{\partial\bm{z}_i}\left(
    \frac{1}{D_{jj}}\sum\limits_{k\in\mathcal{N}(j)}a_{jk}
    \ \dotsi\ \frac{1}{D_{mm}}\sum\limits_{o\in\mathcal{N}(m)}a_{mo}\bm{z}_o
    \right) \notag \\
    &=\frac{\partial}{\partial\bm{z}_i}\left(\sum\limits_{k=1}^l\sum\limits_{ \mathcal{P}^{j\rightarrow i}_k}\prod_{o=k}^1\tilde{a}_{v_{(o-1)},v_{(o)}}\bm{z}_i \right) \notag \\ &=\frac{\partial}{\partial\bm{z}_i}\left(
    \sum\limits_{p=1}^q\prod_{o=k_p}^1\tilde{a}_{v^{(p)}_{(o-1)},v^{(p)}_{(o)}}\bm{z}_i \right).
\end{align}
Here, we substitute the term $\bm{z}_j$ by expansion based on the GCN propagation rule. $\mathcal{P}^{j\rightarrow i}_k$ is a path $[v_{(k)}, v_{(k-1)},\dotsc,v_{(0)}]$ of length $k$ from node $v_j$ to $v_i$, where $v_{(k)}=v_j$ and $v_{(0)}=v_i$. Moreover,  $v_{(o-1)}\in\mathcal{N}\left(v_{(o)}\right)$ for $o=k,k-1,\dotsc,1$. $\tilde{a}_{v_{(o-1)},v_{(o)}}$ denotes the normalized edge weight between $v^{(p)}_{(o-1)}$ and $v^{(p)}_{(o)}$. $k_p$ is the length of the $p$-th path, and we denote node $v_{(o)}$ in the $p$-th path as $v^{(p)}_{(o)}$.
In this expansion, we first aggregate all paths from $v_j$ to $v_i$ of different lengths (with a maximum length of $l$) and ignore other paths that end at other nodes. This is because we only consider the gradient between $v_j$ and $v_i$. In other words, the derivative on other paths will be 0. Denoting the total number of such paths as $q^{(j,i)}$, we can further extract $\partial\bm{z}_i/\partial\bm{z}_i$ as follows:
\begin{equation}
    \begin{aligned}
        \frac{\partial \bm{z}_j}{\partial \bm{z}_i}
    &=\frac{\partial\bm{z}_i}{\partial\bm{z}_i}\cdot\left(
    \sum\limits_{p=1}^{q^{(j,i)}}\prod_{o=k_p}^1\tilde{a}_{v^{(p)}_{(o-1)},v^{(p)}_{(o)}}\right) \\ &=\mathbf{I}\cdot\left(
    \sum\limits_{p=1}^{q^{(j,i)}}\prod_{o=k_p}^1\tilde{a}_{v^{(p)}_{(o-1)},v^{(p)}_{(o)}}\right).
    \end{aligned}
\end{equation}
Therefore, we have
\begin{equation}
\begin{aligned}
        \frac{\|\bm{z}_j^{\star}-\bm{z}_j\|_2}{\|\bm{z}_j\|_2}
        &\approx\|\frac{\partial\bm{z}_j}{\partial\bm{z}_i}\bm{z}_i\|_2\cdot\frac{1}{\|\bm{z}_j\|_2}
     \\ & =\left(
    \sum\limits_{p=1}^{q^{(j,i)}}\prod_{o=k_p}^1\tilde{a}_{v^{(p)}_{(o-1)},v^{(p)}_{(o)}}\right)
    \frac{\|\bm{z}_i\|_2}{\|\bm{z}_j\|_2}.
    \end{aligned}
    \label{eq:distance}
    \end{equation}
In the above equations, the value of $\|\bm{z}_i\|_2/\|\bm{z}_j\|_2$ becomes 1 since all norms are the same. Then we select the path with the largest value of $\prod_{o=k}^1\tilde{a}_{v_{(o-1)},v_{(o)}}$ from all $q^{(j,i)}$ possible paths starting from $v_j$ to $v_i$, denoted as path $p_*$:   
\begin{align}  
    \sum\limits_{p=1}^{q^{(j,i)}}\prod_{o=k_p}^1\tilde{a}_{v^{(p)}_{(o-1)},v^{(p)}_{(o)}}
    &\leq q^{(j,i)} \cdot\max\left(
\prod_{o=k_1}^1\tilde{a}_{v^{(1)}_{(o-1)},v^{(1)}_{(o)}}, 
\ \right.  \notag \\
& \;\;\;\; \left.\dotsi,\ \prod_{o=k_m}^1\tilde{a}_{v^{(q^{(j,i)})}_{(o-1)},v^{(q^{(j,i)})}_{(o)}}
    \right) \notag \\
    &= q^{(j,i)} \cdot\left(
\prod_{o=k_{p_*}}^1\tilde{a}_{v^{(p_*)}_{(o-1)},v^{(p_*)}_{(o)}}
    \right).
\end{align}
Here $k_{p_*}$ is the length of path $p_*$. Since the edge weight is 1 on the path, we can remove the expression of $a$ as follows:
 \begin{equation}
\begin{aligned}  
    \sum\limits_{p=1}^{q^{(j,i)}}\prod_{o=k_p}^1\tilde{a}_{v^{(p)}_{(o-1)},v^{(p)}_{(o)}}
    &\leq q^{(j,i)}\cdot \left(
\prod_{o=k_{p_*}}^1\frac{1}{D^{(p_*)}_{(o),(o)}}
    \right) \\ &=q^{(j,i)}\cdot\left(
    \sqrt[k_{p_*}]{\prod_{o=k_{p_*}}^1D^{(p_*)}_{(o),(o)}}
    \right)^{-k_{p_*}}.
        \end{aligned}
\end{equation} 
Combining with Eq. (\ref{eq:distance}), we can obtain
  \begin{equation}
\begin{aligned}  
\frac{\|\bm{z}_j^{\star}-\bm{z}_j\|_2}{\|\bm{z}_j\|_2}
&\approx \left(\sum\limits_{p=1}^{q^{(j,i)}}\prod_{o=k_p}^1\tilde{a}_{v^{(p)}_{(o-1)},v^{(p)}_{(o)}}\right)
\\ & \leq q^{(j,i)}/(d_{min}^{(j,i)})^{k_{p_*}}
\\ & \leq q^{(j,i)}/(d_{min}^{(j,i)})^{h^{(j,i)}},
        \end{aligned}
\end{equation} 
where $d_{min}^{(j,i)}$ is the geometric mean of node degrees on path $p_*$. Then, we can further utilize $h^{(j,i)}$, which is the shortest path between $v_j$ and $v_i$, to obtain the inequality. 
It is worth mentioning that since we do not explicitly restrict the values of edge weights, our proof can be generalized to other GNNs by choosing appropriate values for the edge weights.
\end{proof}

\textbf{Theorem 3.}
\textit{
Denote $I_{\Gamma}(\tilde{\mathcal{V}})$ as the summation of influence scores to the PDD for node set $\tilde{\mathcal{V}}$. Given a node set $\tilde{\mathcal{V}} \subseteq \mathcal{V}'$ ($|\tilde{\mathcal{V}}| = k$), if $\mathcal{V}_{x}' \cap \mathcal{V}_{y}' = \varnothing $ holds for any pair of nodes $v_{x}, v_{y} \in \tilde{\mathcal{V}}$ ($x \neq y$), then $I_{\Gamma}(\tilde{\mathcal{V}}) = \sum_{v_i \in \tilde{\mathcal{V}}} I_{\Gamma}(v_{i})$.}
\begin{proof}
Recall that 
\begin{align}
\hat{\boldsymbol{W}}_{\epsilon, v_{i}} &\stackrel{\text { def }}{=} \arg \min _{\boldsymbol{W}} L_{\mathcal{V}^{\prime}}(\mathcal{G}, \boldsymbol{W}) \notag \\ & -\epsilon\left(L_{v_{i}}\left(\mathcal{G}_{i}, \boldsymbol{W}\right)+\tilde{L}_{\mathcal{V}_{i}^{\prime}}\left(\mathcal{G}_{i}, \boldsymbol{W}\right)\right).
\end{align}
Considering the computation graphs of any two nodes in $\tilde{\mathcal{V}}$ do not overlap with each other, each node $v_i$ in $\tilde{\mathcal{V}}$ corresponds to a term of $L_{v_{i}}\left(\mathcal{G}_{i}, \boldsymbol{W}\right)+\tilde{L}_{\mathcal{V}_{i}^{\prime}}\left(\mathcal{G}_{i}, \boldsymbol{W}\right)$ that is non-dependent on other training nodes (or the training nodes in their computation graphs) to be deleted.
Therefore, the nodes in set $\tilde{\mathcal{V}}$ correspond to a linear sum of the $L_{v_{i}}\left(\mathcal{G}_{i}, \boldsymbol{W}\right)+\tilde{L}_{\mathcal{V}_{i}^{\prime}}\left(\mathcal{G}_{i}, \boldsymbol{W}\right)$ term. Consequently, we have
\begin{align}
I_{G}(\tilde{\mathcal{V}}) &=  -\left(\frac{\partial G}{\partial \bm{\hat{W}}_{\epsilon, \tilde{\mathcal{V}}}} \right)^{\top} \cdot \left(\frac{\partial^{2} L_{\mathcal{V}'}(\mathcal{G},\bm{W})}{\partial\bm{W}^{2}}\right)^{-1} \notag \\ &\;\;\;\; \cdot \left( \sum_{k=1}^{K} \frac{\partial L_{v_{i_k}}\left(\mathcal{G}_{i_k},\bm{W}\right)}{\partial\bm{W}} + \sum_{k=1}^{K} \frac{\partial \tilde{L}_{\mathcal{V}_{i_k}'}(\mathcal{G}_{i_k},\bm{W})}{\partial\bm{W}} \right) \notag  \\ 
&= - \sum_{k=1}^{K} \left(\frac{\partial G}{\partial \bm{\hat{W}}_{\epsilon, \tilde{\mathcal{V}}}} \right)^{\top} \cdot \left(\frac{\partial^{2} L_{\mathcal{V}'}(\mathcal{G},\bm{W})}{\partial\bm{W}^{2}}\right)^{-1} \notag \\ &\;\;\;\; \cdot  \left( \frac{\partial L_{v_{i_k}}\left(\mathcal{G}_{i_k},\bm{W}\right)}{\partial\bm{W}} +  \frac{\partial \tilde{L}_{\mathcal{V}_{i_k}'}(\mathcal{G}_{i_k},\bm{W})}{\partial\bm{W}} \right)   \notag  \\
&=  \sum_{k=1}^{K} I_{G}(v_{i_k}).
\end{align}
\end{proof}


\end{document}